\documentclass[10pt,journal]{IEEEtran}
\usepackage{amsmath,amssymb,amsfonts}
\usepackage{pifont}
\usepackage{algorithmicx}
\usepackage[noend,ruled, linesnumbered, vlined]{algorithm2e}
\usepackage{graphicx}
\usepackage{epstopdf}
\usepackage{textcomp}
\usepackage{xcolor}
\usepackage{bm}
\usepackage{array}
\usepackage{ntheorem}
\usepackage{subfigure} 
\usepackage{cases}
\usepackage{stfloats}
\usepackage{url}
\usepackage{verbatim}
\usepackage{algpseudocode}
\usepackage{enumerate}
\usepackage{ulem}
\usepackage{booktabs}
\usepackage{cite}

\usepackage{color}

\usepackage{enumitem}[topsep=3pt,itemsep=3pt]

\allowdisplaybreaks[4]
\def\BibTeX{{\rm B\kern-.05em{\sc i\kern-.025em b}\kern-.08em
    T\kern-.1667em\lower.7ex\hbox{E}\kern-.125emX}}
\definecolor{b}{rgb}{0.0, 0, 0}
\definecolor{r}{rgb}{0, 0, 0}

\usepackage{tikz,xcolor,hyperref}
\definecolor{lime}{HTML}{A6CE39}
\DeclareRobustCommand{\orcidicon}{
    \begin{tikzpicture}
    \draw[lime, fill=lime] (0,0) 
    circle [radius=0.16] 
    node[white] {{\fontfamily{qag}\selectfont \tiny ID}};    \draw[white, fill=white] (-0.0625,0.095) 
    circle [radius=0.007];    \end{tikzpicture}
    \hspace{-2mm}}
\foreach \x in {A, ..., Z}{%
    \expandafter\xdef\csname orcid\x\endcsname{\noexpand\href{https://orcid.org/\csname orcidauthor\x\endcsname}{\noexpand\orcidicon}}
    }

\theoremseparator{.}
\newskip\theorempreskipamount
\newskip\theorempostskipamount

\theorembodyfont{}

\begin{document}

\title{Joint Resource Management for Energy-efficient UAV-assisted SWIPT-MEC: A Deep Reinforcement Learning Approach}

\author{Yue Chen,
Hui Kang,
Jiahui Li, 
Geng Sun, \IEEEmembership{Senior Member, IEEE},
Boxiong Wang,
Jiacheng Wang,
Cong Liang,
Shuang Liang,
and Dusit Niyato,~\IEEEmembership{Fellow, IEEE}


\thanks{This work is supported in part by the National Natural Science Foundation of China (62172186, 62272194, 62471200), in part by the Science and Technology Development Plan Project of Jilin Province (20240302079GX, 20250102210JC), in part by the Postdoctoral Fellowship Program of China Postdoctoral Science Foundation (GZC20240592), in part by the China Postdoctoral Science Foundation General Fund (2024M761123), and in part by the Scientific Research Project of Jilin Provincial Department of Education (JJKH20250117KJ). \textit{(Corresponding authors: Jiahui Li and Geng Sun.)}
\par Yue Chen, Hui Kang, Jiahui Li, and Boxiong Wang are with the College of Computer Science and Technology, Jilin University, Changchun 130012, China. Hui Kang is also with the Key Laboratory of Symbolic Computation and Knowledge Engineering of Ministry of Education, Jilin University, Changchun 130012, China. (E-mails: yuechen23@mails.jlu.edu.cn; kanghui@jlu.edu.cn; lijiahui@jlu.edu.cn; wangbx0320@163.com).
\par Geng Sun is with the College of Computer Science and Technology, Jilin University, Changchun 130012, China, and also with the Key Laboratory of Symbolic Computation and Knowledge Engineering of Ministry of Education, Jilin University, Changchun 130012, China. He is also with the College of Computing and Data Science, Nanyang Technological University, Singapore 639798 (E-mail: sungeng@jlu.edu.cn).
\par Jiacheng Wang and Dusit Niyato are with the College of Computing and Data Science, Nanyang Technological University, Singapore (E-mails: jiacheng.wang@ntu.edu.sg; dniyato@ntu.edu.sg).
\par Cong Liang is with China Mobile Communications Group Jilin Co., Ltd. (E-mails: liangcong@jl.chinamobile.com).
\par Shuang Liang is with the School of Information Science and Technology, Northeast Normal University, Changchun 130117, China (E-mails: liangshuang@nenu.edu.cn).
}
}

\IEEEtitleabstractindextext{
\begin{abstract}
The integration of simultaneous wireless information and power transfer (SWIPT) technology in 6G Internet of Things (IoT) networks faces significant challenges in remote areas and disaster scenarios where ground infrastructure is unavailable. This paper proposes a novel unmanned aerial vehicle (UAV)-assisted mobile edge computing (MEC) system enhanced by directional antennas to provide both computational resources and energy support for ground IoT terminals. However, such systems require multiple trade-off policies to balance UAV energy consumption, terminal battery levels, and computational resource allocation under various constraints, including limited UAV battery capacity, non-linear energy harvesting characteristics, and dynamic task arrivals. To address these challenges comprehensively, we formulate a bi-objective optimization problem that simultaneously considers system energy efficiency and terminal battery sustainability. We then reformulate this non-convex problem with a hybrid solution space as a Markov decision process (MDP) and propose an improved soft actor-critic (SAC) algorithm with an action simplification mechanism to enhance its convergence and generalization capabilities. Simulation results have demonstrated that our proposed approach outperforms various baselines in different scenarios, achieving efficient energy management while maintaining high computational performance. Furthermore, our method shows strong generalization ability across different scenarios, particularly in complex environments, validating the effectiveness of our designed boundary penalty and charging reward mechanisms.
\end{abstract}
        }

\maketitle
\IEEEdisplaynontitleabstractindextext
\IEEEpeerreviewmaketitle

\section{Introduction}
\label{sec:Introduction}

\par \IEEEPARstart{T}{he} simultaneous wireless information and power transfer (SWIPT) has emerged as a promising technology for Internet of Things (IoT) applications in sixth-generation (6G) wireless networks~\cite{Pan2023, VuNK22, SuFTCFZW23, ZhouHYS22}. The anticipated rapid expansion of IoT networks, characterized by large-scale deployment, automation, and low power consumption~\cite{Li2025, NguyenDPSLNDP22}, presents significant challenges in providing a reliable power supply and network connectivity. Specifically, traditional approaches such as wired connections and battery replacements incur substantial maintenance costs~\cite{ChoiHAJKKK20}, particularly problematic given the increasing computational demands of IoT terminals requiring continuous network connectivity. To address these challenges, SWIPT enables simultaneous energy and information transmission via radio frequency (RF) signals from ground-based stations. However, this solution exhibits limitations in remote areas or disaster scenarios where ground infrastructure is unavailable or damaged, potentially compromising network coverage and communication performance.

\par Low-altitude unmanned aerial vehicles (UAV) emerge as an effective solution to challenges in traditional IoT networks and SWIPT implementation. Functioning as mobile base stations, UAVs offer cost-effectiveness, mobility, and ease of deployment~\cite{JiangXYFZL22, Qu2024}, which attributes are critical for dynamic IoT environments. Their ability to adjust position in real-time establishes line-of-sight (LoS) connections with ground IoT terminals, enhancing wireless network performance and robustness~\cite{WangG19}. Additionally, UAVs equipped with mobile edge computing (MEC) servers bring computational resources closer to resource-constrained IoT terminals, reducing processing latency and improving operational efficiency~\cite{Sun2024c, Yan2024}. By simultaneously supporting SWIPT and MEC functionalities, UAVs effectively address both energy harvesting (EH) and computational demands in dynamic and resource-constrained IoT environments.

\par However, integrating MEC and SWIPT into UAV-assisted 6G-enabled IoT networks while efficiently managing computational tasks presents several challenges~\cite{Wang2024c, Wang2024d, Wang2025b}. \textit{First}, UAVs face inherent energy limitations due to battery size and weight constraints~\cite{Wang2020}, which necessitate efficient energy allocations for movement, communication, and computation. In particular, the decision variables of UAV flight strategy, such as velocity, have a nonlinear relationship with propulsion power consumption, thus further posing optimization difficulty~\cite{Sun2021}. \textit{Second}, the limited computational resources and communication bandwidth of UAVs may be insufficient to meet the computational and energy demands of multiple IoT devices simultaneously~\cite{Wang2025earlyaccess}, which means that the energy charging fairness among these IoT devices must be considered. \textit{Third}, stochastic task arrivals and channel condition fluctuations bring dynamics to the considered UAV-assisted multi-terminal SWIPT-MEC network, causing the static optimization method to lack adaptability~\cite{Sun2024,Sun2025earlyaccess}.

\par To address these challenges comprehensively, we formulate a bi-objective optimization problem related to the total energy consumption of the system and the average battery level of the terminals in a UAV-assisted multi-terminal SWIPT-MEC system. The main contributions of this paper are summarized as follows:

\begin{itemize}
    \item \textit{UAV-assisted Multi-terminal SWIPT-MEC System with directional antenna:} We consider a novel UAV-assisted network that integrates SWIPT with MEC support for multiple terminals, specifically designed for regions without coverage from ground base stations. In this system, the UAV functions as a mobile base station providing MEC services while simultaneously employing SWIPT to deliver wireless charging support to terminals. Simultaneously, we consider the UAV with directional antennas to enhance the downlink transmission signal quality. It will improve both communication efficiency and charging capabilities, which ultimately extends the average battery runtime of terminals.

    \item \textit{Formulation of a Bi-objective Optimization Problem:} We formulate a bi-objective optimization problem, aiming to minimize the total energy consumption of the system and maximize the average battery level of the terminals while ensuring charging fairness among terminals simultaneously. The formulated problem is a non-convex mixed-integer nonlinear programming problem with a hybrid solution space, i.e., discrete and continuous variables.

    \item \textit{DRL-base Solution:} We propose a novel deep reinforcement learning (DRL)-based off-policy optimization approach. We first reformulate the optimization problem as a Markov decision process (MDP). Subsequently, we design an action simplification mechanism to address the hybrid action space. Based on this, we propose an improved soft actor-critic (SAC) algorithm that enhances the processing and modeling capabilities of neural networks, thereby resulting in superior convergence performance.

    \item \textit{Performance Evaluation and Analysis:} Simulation results demonstrate that the proposed algorithm outperforms various baselines across key metrics, yielding a 47.86\% improvement in average terminal retained energy and a 65.15\% enhancement in charging fairness. Moreover, the proposed algorithm exhibits fast convergence and effectively learns the maximum entropy optimal policy, successfully balancing exploration and exploitation in complex decision spaces. Furthermore, it demonstrates strong generalization capabilities across diverse terminal distributions, particularly in the case of uneven distribution.
\end{itemize}

\par The rest of this paper is organized as follows. Section~\ref{sec:Related Work} reviews the related research activities. Section~\ref{sec:System Model and Problem Formulation} presents the directional antenna-enhanced UAV-assisted multi-terminal SWIPT-MEC network and optimization problem formulation. Section~\ref{sec:DRL-Based Method} proposes our DRL-based solution. Simulation results are presented and analyzed in Section~\ref{sec:Simulation And Analyses}. Finally, this work is concluded in Section~\ref{sec:Conclusion}.

%
%
\section{Related Work}
\label{sec:Related Work}

\par In this section, we review related works on paradigms, optimization objectives, and optimization methods in UAV-assisted SWIPT-MEC networks.

%
%
\subsection{Paradigms in MEC Networks}
\label{subsec:Paradigms in MEC Networks}

\par MEC technology allows computation-intensive and latency-critical applications to be offloaded from resource-constrained mobile devices to proximate network edges~\cite{Sun2025}, which attracts considerable research interest. The related literature on MEC networks can be classified into three distinct research paradigms.

%
%
\textit{Ground MEC Networks:} Offloading computational tasks from mobile devices to ground-based stations represents the foundational paradigm of MEC. For example, the authors in~\cite{Widiyanti2023} proposed an Internet of Video Things (IoVT) network based on MEC, combined with the reconfigurable intelligent surface (RIS). However, ground-based MEC systems face significant challenges related to scalability and disaster recovery capabilities. In remote areas, these systems often become inoperative due to insufficient base station coverage and inadequate infrastructure support.

%
%
\textit{UAV-assisted MEC Networks:} The integration of UAVs effectively addresses critical limitations of traditional ground MEC networks by deploying UAVs as mobile base stations or communication relays. For instance, the authors in~\cite{Song2023} utilized a UAV as either an MEC server or wireless relay within a UAV-assisted MEC system. While these contributions advance UAV-assisted MEC capabilities, they overlooked the energy requirements of mobile terminals, which are particularly crucial for energy-constrained devices such as sensors.


%
%
\textit{UAV-assisted WPT-MEC Networks:} Some studies have now applied wireless power transmission (WPT) technology, utilizing UAVs as power sources to charge mobile devices. The authors in~\cite{Zeng2023} investigated a UAV-assisted WPT-MEC system, where the UAV provided both power and computation services to mobile terminals.

\par Despite advances in the field, research remains limited on utilizing UAVs to simultaneously provide charging and downlink information transmission services for terminals in MEC networks. This capability is particularly valuable in areas lacking ground-based infrastructure and better aligns with practical implementation demands.

%
%
\subsection{Optimization Objectives in UAV-assisted Networks with WPT/SWIPT}
\label{Optimization Objectives in UAV-assisted Networks with WPT/SWIPT}

\par In the paradigms of UAV-assisted WPT/SWIPT networks, common optimization objectives are typically categorized into the following two main aspects.

%
%
\textit{Energy-related Objectives:} Several studies focus on energy-related objectives, such as energy consumption and energy efficiency (EE). For example, the authors in~\cite{Du0WZZC19} proposed a novel time division multiple access (TDMA)-based workflow model, aiming to minimize the total energy consumption of the UAV. Likewise, the authors in~\cite{Liu2020} formulated an optimization problem in the scenario of a UAV-enabled wireless-powered cooperative MEC system, seeking to minimize the total required energy of the UAV. The authors in~\cite{SuFTCFZW23} investigated the EE maximization optimization problem for device-to-device (D2D) communications underlaying UAVs-assisted industrial IoT networks with SWIPT. Nonetheless, the aforementioned studies concentrate on the energy consumption or EE of UAVs or systems, without considering optimization objectives related to the performance of ground terminals.

%
%
\textit{Terminals-related Objectives:} Several studies focus on the objectives related to terminal performance. For instance, the authors in~\cite{Liu2024} considered a heterogeneous MEC system with multiple energy-limited IoT devices and a UAV to maximize the minimum task computation data volume among all active devices. Similarly, the authors in~\cite{Baduge2024} integrated SWIPT-enabled non-orthogonal multiple access (NOMA) with a UAV, aiming to maximize achievable rates for the ground users. The authors in~\cite{Heo2024} considered a UAV-assisted SWIPT system to maximize the sum of the logarithmic average throughput of the GNs. However, the aforementioned research emphasizes the terminal-related metrics but neglects the overall system performance optimization.

\par As a result, we construct a directional antenna-enhanced UAV-assisted multi-terminal SWIPT-MEC system. In contrast to existing studies, our approach targets optimization objectives at two levels. At the macro level, we aim to minimize UAV energy consumption, while at the micro level, our objective is to maximize the battery energy of the terminals, ensuring charging fairness among the terminals.

%
%
\subsection{Optimization Methods in UAV-assisted Networks with WPT/SWIPT}
\label{Optimization Methods in UAV-assisted with Networks WPT/SWIPT}

\par To achieve the aforementioned optimization objectives, various methods have been employed in the literature, which can be broadly categorized into the following categories.

%
%
\textit{Convex optimization methods:} Several studies use traditional static convex or non-convex optimization methods. For example, the authors in~\cite{Jalali2024} investigated a UAV-assisted SWIPT network and proposed a convex optimization method that combines successive convex approximation (SCA) and a quadratic transformation approach. However, convex optimization methods have limitations in terms of applicability and global convergence, especially in weakly constrained problems involving complex structures and multiple local optima.


%
%
\textit{Evolutionary Computation Methods:} Several studies employ evolutionary computation methods to address non-convex optimization problems due to the coupled variables. For instance, the authors in~\cite{Feng2020} proposed a multi-objective evolutionary algorithm based on decomposition (MOEA/D) for an emergency communication framework in UAV-enabled SWIPT IoT networks. However, evolutionary computation methods are prone to being trapped in local optima when confronted with high-dimensional data and multimodal problems, and their performance is often highly sensitive to parameter settings.

%
%
\textit{Traditional DRL Methods:} More recently, some studies have attempted to utilize conventional DRL methods to solve dynamic nonlinear programming problems and address the complexities of dynamic environments to achieve enhanced system performance. Specifically, the authors in~\cite{Shi2024} considered a computation-intensive MEC network based on NOMA-SWIPT and proposed a multi-agent deep deterministic policy gradient (MADDPG)-based resource management algorithm. The authors in~\cite{Chhea2024earlyaccess} studied an energy-efficient UAV network enhanced by RIS with SWIPT, maximizing the average EE by employing the deep Q-learning (DQL) framework. However, the aforementioned works do not offer targeted improvements to address the complexities and variability of dynamic environments.


\par In summary, distinct from prior studies, we propose a novel DRL-based method for directional antenna-enhanced UAV-assisted multi-terminal SWIPT-MEC systems. This approach effectively tackles dynamic task offloading and UAV scheduling challenges while addressing high-dimensional action-state spaces with a low-complexity solution. Consequently, our method achieves an optimal strategy that balances competing optimization objectives.

%
%
\section{System Model and Problem Formulation}
\label{sec:System Model and Problem Formulation}

\par In this section, we introduce a directional antenna-enhanced UAV-assisted multi-terminal SWIPT-MEC system with details presented as follows.

\par
\begin{figure}[t] 
	\centering
	\setlength{\abovecaptionskip}{2pt}%
	\setlength{\belowcaptionskip}{2pt}%
	\includegraphics[width =3.5in]{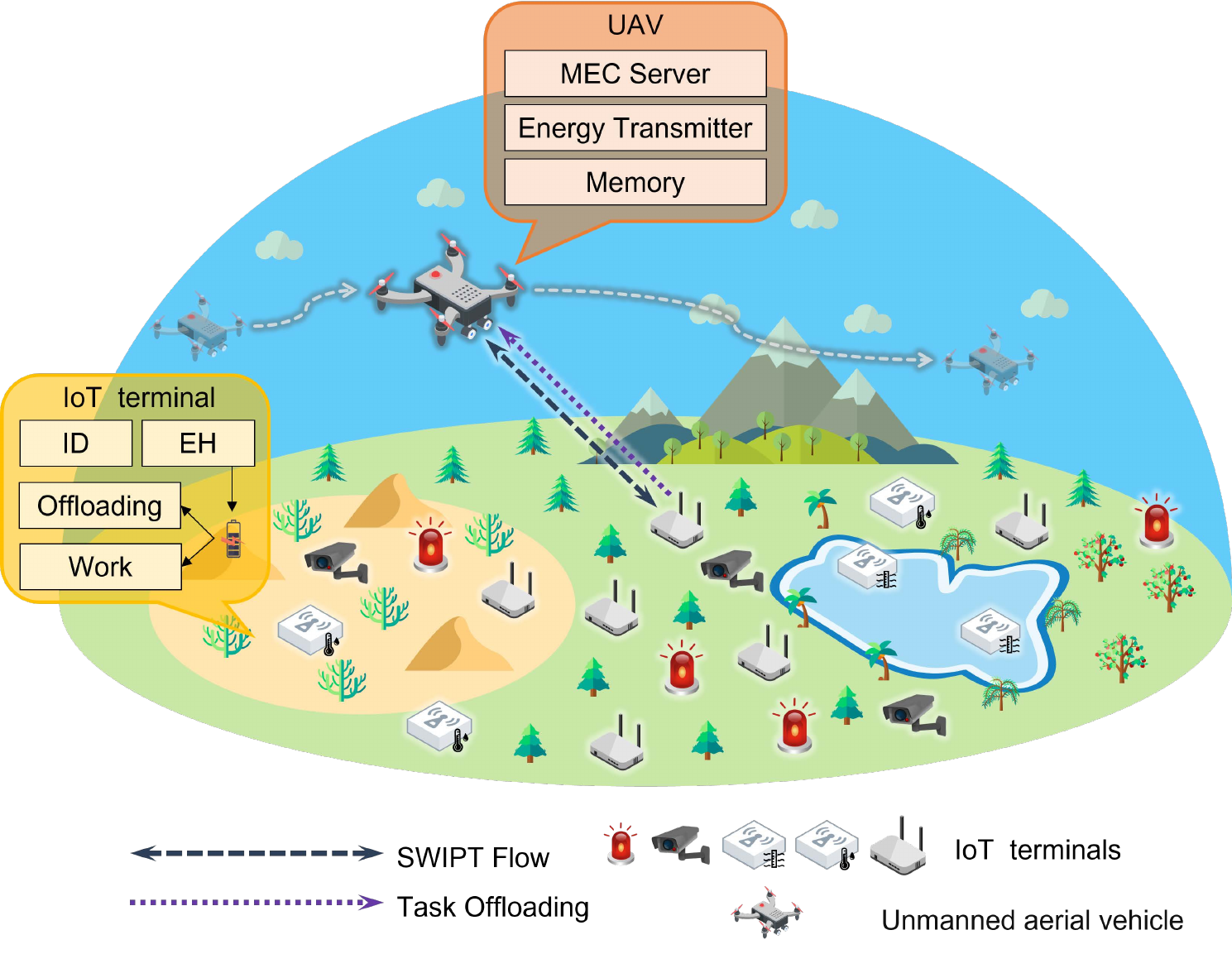}
	\caption{The directional antenna-enhanced UAV-assisted multi-terminal SWIPT-MEC system. }
	\label{SystemModel}
	\vspace{1em}
\end{figure}

 Fig.~\ref{SystemModel} illustrates our proposed system comprising fixed ground IoT terminals, denoted as $i \in \mathcal{I} = \{1, \ldots, I\}$. These terminals need to perform computational tasks and transfer information simultaneously for applications such as water quality detection, fire warning, and humidity monitoring. However, in remote areas, the absence of available ground infrastructure makes it impossible to directly provide service to these terminals~\cite{Sun2024}. Moreover, these terminals are inherently constrained by their limited computational resources and battery capacity, which significantly restricts their task-processing capabilities and service lifetimes.

\par To address these challenges, we deploy a UAV $u$ equipped with a directional antenna to provide communication, computing services, and energy supply through SWIPT technology~\cite{Jiang2022}. Note that we consider one UAV for the sake of simplicity, and the proposed model can be further extended to multi-UAV cases. Specifically, we first divide the network into clusters using algorithms such as k-means and then apply our modeling approach to each cluster independently.

\par Fig.~\ref{MECPhase} illustrates our discrete-time model that partitions continuous time into $T$ discrete equal-duration time slots. Specifically, we define the collection of time slots as $\mathcal{T} = \{1, \ldots, t, \ldots, T\}$, and the duration of each time slot is denoted as $\tau$, which is chosen to be sufficiently small such that each time slot maintains quasi-static conditions. Thus, we consider that the channel state information (CSI) and the position of the UAV remain constant within a single time slot~\cite{ZhouHYS22}. When an IoT terminal $i$ offloads its task to the UAV for processing, each time slot is composed of the following phases:

\begin{itemize}
    \item \textit{Task Offloading Phase}: The IoT terminal offloads its computing task to the UAV.
    
    \item \textit{SWIPT and Computation Phase}: After receiving the task, the UAV processes it and transmits the signal to the terminal via a directional antenna. These signals simultaneously deliver both information (external data and computation results) and energy to the terminal. The terminal then performs signal splitting to enable both information decoding (ID) and EH.
\end{itemize}

\par Without loss of generality, we denote the coordinates of the ground IoT terminal $i$ as $p_i = (x_i, y_i, 0)$. The UAV operates at a fixed altitude $H$~\cite{SuFTCFZW23, ZhouHYS22}, with its position at time slot $t$ represented as $p_u^t = (x_u^t, y_u^t, H)$.

%
%
\subsection{Task Model}
\label{subsec:Task Model}

\par In this section, we describe the task generation model and the associated offloading decision-making process.

\par For the downlink communication, terminal $i$ receives information data $D_{i,r}$ from the UAV, which possesses sufficient storage capacity to carry all required information.

\begin{figure}[t] 
	\centering
	\setlength{\abovecaptionskip}{2pt}%
	\setlength{\belowcaptionskip}{2pt}%
	\includegraphics[width =3.5in]{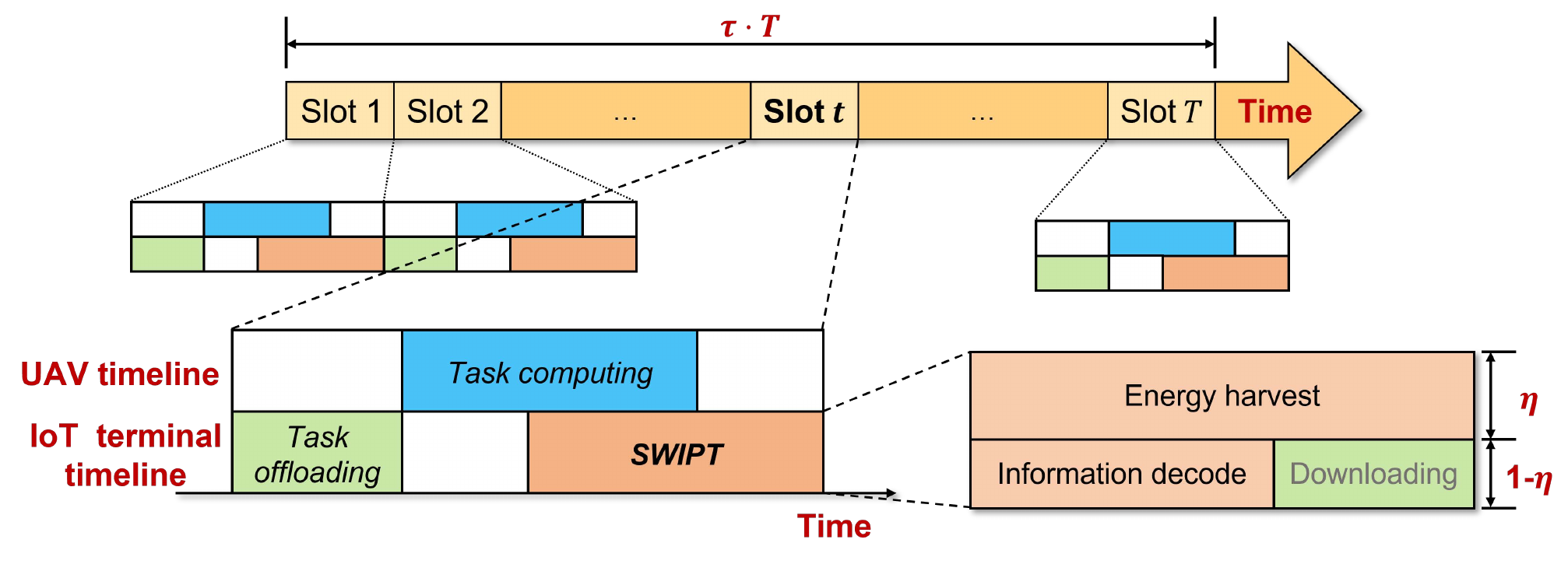}
	\caption{The time slot division model in the offloading scenario.}
	\label{MECPhase}
	\vspace{1em}
\end{figure}

\par For the uplink communication, computational tasks at IoT terminal $i$ arrive according to a Bernoulli process during time slot $t$~\cite{Meng2024}, generating either 0 or 1 task in a time slot. Each task is characterized by a tuple $\{D_{i,p}^t, t_{\text{gen}}, C_i\}$, where $D_{i,p}^t$ represents the data size of task (bit), $t_{\text{gen}}$ indicates the generation time slot, and $C_i$ denotes the computation intensity (cycles/bit). We consider these computational tasks to be atomic, requiring completion within a single time slot. Consequently, only two processing strategies are available: local computation or complete offloading to the UAV~\cite{Wang2022}. This approach allows complex tasks comprising multiple indivisible units to be treated as separate atomic tasks, enhancing problem tractability.

\par To effectively describe task processing strategies of terminal $i$, we define a binary indicator $o_i^t \in \{0,1\}$ representing the offloading decision in time slot $t$, where $o_i^t=1$ indicates task offloading and $o_i^t=0$ represents local processing. The selection of $o_i^t$ depends on the size of the task and processing requirements, as well as current communication conditions and resource availability. As illustrated in Fig.~\ref{MECPhase}, the time slot division model can be expressed as follows:
\begin{equation}
    \label{MECTimeConstrain}
    \tau_{\text{up}} + \tau_{\text{S}} \leq \tau,
\end{equation}

\noindent where $\tau_{\text{up}} = \frac{o_i^t\left(D_{i,p}^t+\delta_{i\to u}\right)}{R_{i\to u}}$ represents the task offloading time. The SWIPT time $\tau_{\text{S}}$ is calculated as the maximum of UAV computing time, EH time, and ID time: $\tau_{\text{S}} = \max\left\{\frac{o_i^tC_iD^t_{i,p}}{f_u}, \min\left(\frac{E_{\text{max}}-E_i}{E_{u\to i}}, \tau-\tau_{\text{up}}\right), \frac{D_{i,r}+\delta_{u\to i}}{R_{u\to i}}\right\}$. Here, $f_u$ is the CPU frequency (cycles/s) of the UAV, $\delta_{i\to u}$ and $\delta_{u\to i}$ are the protocol overheads required for uplink and downlink, respectively. Additionally, $E_{\text{max}}$ represents the maximum battery capacity, $E_i$ denotes the remaining battery level of terminal $i$, while $E_{u \rightarrow i}$ and $R_{u \rightarrow i}$ represent the EH rate and achievable downlink data rate, respectively, which will be detailed in the next section.

%
%
\subsection{SWIPT Model}
\label{subsec:SWIPT Model}

\par In this part, we introduce the SWIPT model, including the air-to-ground channel models and the non-linear EH model.

%
%
\subsubsection{Air-to-ground Channel Model}
\label{subsec:Air-to-ground Channel Model}

\par We employ a probabilistic LoS channel model for communication, where the LoS probability $P_\mathrm{LoS}$ is given by~\cite{Yaliniz2016}: $P_{\mathrm{LoS}} = \frac{1}{1 + a_1 \exp\left(-b_1\left(\frac{180}{\pi}\arctan\left(\frac{H}{d_i^t}\right) - a_1\right)\right)}$. Here, parameters $a_1$ and $b_1$ are environment-dependent, $H$ represents the UAV altitude, and $d_i^t$ denotes the horizontal distance between the UAV and terminal $i$ at time slot $t$, which is calculated as $d_i^t = \sqrt{\left(x_u^t - x_i\right)^2 + \left(y_u^t - y_i\right)^2}$.

\par Based on $P_{\mathrm{LoS}}$, the path loss between the UAV and terminal $i$ for a given time slot $t$ can be expressed as follows:
\begin{equation}
\label{PathLoss}
    L_i^t = 20 \log_{10}\left(\frac{4\pi f_c\|p_u^t,p_i\|}{c}\right) + P_{\mathrm{LoS}}\eta_{\mathrm{LoS}} + (1-P_{\mathrm{LoS}})\eta_{\mathrm{NLoS}},
\end{equation}

\noindent where $\|\cdot\|$ denotes the Euclidean norm, $f_c$ is the carrier frequency, and $c$ is the speed of light. The terms $\eta_\mathrm{LoS}$ and $\eta_\mathrm{NLoS}$ represent the additional losses associated with LoS and NLoS links, respectively, which are derived from free-space path loss and depend on specific environmental conditions. Consequently, the corresponding channel gain can then be calculated as $h^t_i=10^{{-L_i^t}/{10}}$

\par Furthermore, to mitigate the impact of channel fading on transmission performance, we consider that the UAV is equipped with a directional antenna, similar to~\cite{Jiang2022, Zeng2019}. Accordingly, the antenna gain can be expressed approximately as follows:
\begin{equation}
    \label{AntennaGain}
    G_i^t=
    \begin{cases}
        \frac{2.28}{\beta^2}, & d_i^t \leq H\tan(\beta)\\
        0, & \text{otherwise}
    \end{cases}.
\end{equation}

\noindent where $\beta$ represents the half-power beamwidth (rad) of the antenna.

\par In the SWIPT communication framework based on the aforementioned model, PS technology is employed to perform both ID and EH simultaneously. In this case, terminal $i$ divides the received RF signal according to PS ratio $\eta$, allocating a fraction $(1-\eta)$ for ID and the remaining fraction $\eta$ for EH. Therefore, given the total transmit power of the UAV $P_{\text{tran}}$, the signal-to-noise ratio (SNR) at the terminal $i$ can be calculated as follows:
\begin{equation}
\label{SNR}
    \gamma_i = \frac{(1-\eta)h^t_iP_{\text{tran}}G_i^t}{\sigma^2},
\end{equation}

\noindent where $\sigma^2$ denotes the power of the Gaussian white noise.

\par Therefore, according to the Shannon theorem, the achievable data rate for terminal $i$ receiving the information of the UAV is given as follows:
\begin{equation}
\label{R_u2i}
    R_{u\to i} = B \log_2\left(1+\gamma_i\right),
\end{equation}

\noindent where $B$ denotes the total bandwidth. Conversely, when terminal $i$ transmits to the UAV, the achievable data rate is expressed as:
\begin{equation}
\label{R_i2u}
    R_{i\to u} = B \log_2\left(1+\frac{P_ih^t_iG_i^t}{\sigma^2}\right),
\end{equation}

\noindent where $P_i$ represents the transmit power of terminal $i$.

%
%
\subsubsection{Non-linear EH Model}
\label{subsec:Non-linear EH Model}

\par For accurate calculation of terminals' EH rates, we adopt a non-linear EH model based on the Logistic function to characterize energy transmission from UAV to terminal. This model represents actual EH circuit characteristics more precisely than linear models. Specifically, the non-linear EH model is formulated as~\cite{Boshkovska2015}:
\begin{equation}
    \mathbb{F}(P_{in}) = \frac{\frac{P_{max}}{1 + e^{-a_2(P_{in} - b_2)}} - \frac{P_{max}}{1 + e^{a_2b_2}}}{1 - \frac{1}{1 + e^{a_2b_2}}},
\end{equation}

\noindent where $\mathbb{F}(P_{in})$ represents the harvested energy under input power $P_{in}$, the maximum harvested power is denoted as $P_{max}$, and constant parameters $a_2$ and $b_2$ determine the non-linear characteristics related to circuit resistance, capacitance, and diode forward voltage.

\par In summary, the achievable EH rate can be calculated as follows:
\begin{equation}
\label{E_u2i}
    E_{u\to i}(t) = \mathbb{F}\left(\eta P_{tran}h^t_iG_i^t\right).
\end{equation}

%
%
\subsection{Computation Model}
\label{subsec:Computation Model}

\par Within the proposed system, tasks are executed either via local computation or edge computing during their generation time slot.

\par \textit{If the task is executed locally:} Specifically, we consider that the CPU frequency of the terminal $i$, denoted as $f_i$, is sufficient to process tasks within a given time slot. In this context, any task generated by terminal $i$ satisfies the constraint $C_iD^t_{i,p} \leq \tau f_i$. Consequently, similar to~\cite{Sun2024}, the computational energy consumption of the terminal $i$ for a single task can be expressed as follows:
\begin{equation}
\label{E_i-com}
    E_{i-com}(t)=P_{i-com}\cdot \frac{\left(1 - o_i^t\right) C_iD^t_{i,p}}{f_i},
\end{equation}

\noindent where the computational power of the terminal is represented as $P_{i-com}=k(f_i)^{\nu}$~\cite{Jiang2020}, where $k$ is the effective capacitance coefficient, and $\nu$ is a constant. However, when the remaining energy is very low, the terminal may not have enough energy to process the task. To ensure that the terminal has enough energy to maintain normal operation after processing the task, we define a constraint as follows:
\begin{equation}
\label{ConstraintEiCom}
    E_{i-com}(t)<E_i(t)-\left(E_{min}+\delta_e\right), \forall i \in \mathcal{I}, t \in \mathcal{T},
\end{equation}

\noindent where $\delta_e$ represents the reserved energy of a terminal to avoid excessively low energy levels after computation.

\par \textit{If the task is offloaded to the UAV:} With the CPU frequency $f_u$ (cycle/s) of the UAV, the computation energy consumption over flight time $T$ is given as follows:
\begin{equation}
\label{E_u-com}
    E^T_{u-com}=P_{u-com}\sum_{t=1}^T\sum_{i=1}^I\frac{o_i^tC_iD^t_{i,p}}{f_u},
\end{equation}

\noindent where the computational power of the UAV $P_{u-com}=k(f_u)^{\nu}$.

%
%
\subsection{Energy Consumption Model}
\label{subsec:Energy Consumption Model}

\par In this section, we present the energy consumption models for both the UAV and IoT terminals.

%
%
\subsubsection{UAV Energy Consumption}
\label{subsec:UAV Energy Consumption}

\par The total energy consumption of the UAV comprises three main components, which can be calculated as follows:
\begin{equation}
\label{Euav}
    E^T_{uav}=E^T_{u-move}+E^T_{u-tran}+E^T_{u-com},
\end{equation}

\noindent where the first component $E^T_{u-move}$ denotes propulsion energy, calculated as in~\cite{Zeng2019a}. Note that $E^T_{u-move}$ varies quadratically with flight velocity, thereby making flight strategy optimization critical for energy efficiency.

\par The second component $E^T_{u-tran}$ represents the communication energy consumption, which encompasses the energy associated with ID and EH processes. With transmit power $P_{tran}$, this component at time $T$ is expressed as: $E^T_{u-tran}=P_{tran}\sum_{t=1}^T\max\left\{\frac{D_{i,r}+\delta_{u\to i}}{R_{u\to i}},\min\left\{\frac{E_{max}-E_i}{E_{u\to i}},\tau-\tau_{up}\right\}\right\}$. Note that the transmission time in each slot $t$ is determined by the concurrent execution of ID and EH processes.

\par The third component $E^T_{u-com}$ represents the computation energy consumption, calculated according to Eq.~\eqref{E_u-com}.

%
%
\subsubsection{Terminal Energy Consumption}
\label{subsec:Terminal Energy Consumption}

\par The terminal energy consumption can be calculated as follows:
\begin{equation}
\label{Eterminal}
    E^T_{terminal}=\sum_{t=1}^T \sum_{i=1}^I \left(E_{i-com}(t) + E_{i-tran}(t)\right),
\end{equation}

\noindent where the first component $E_{i-com}(t)$ represents the computational energy consumption of terminal $i$, calculated according to Eq.~\eqref{E_i-com}.

\par The second component $E_{i-tran}(t)$, the transmission energy consumption, is incurred when task $\{D_{i,p}^{t}, t_{\mathrm{gen}}, F_{i}\}$ is offloaded. In this case, terminal $i$ utilizes a portion of its local storage energy for uplink data transmission. As indicated by Eq.~\eqref{MECTimeConstrain}, the offloading time is $\tau_{up}$ with terminal transmit power $P_i$, resulting in $E_{i-tran}(t)=P_i \cdot \tau_{up}$. Note that in each time slot, only one type of energy consumption (either $E_{i-com}(t)$ or $E_{i-tran}(t)$) is incurred, depending on whether $o_i^t=0$ or $o_i^t=1$, respectively. Additionally, similar to Eq.~\eqref{ConstraintEiCom}, $E_{i-tran}(t)$ should satisfy the following constraint:
\begin{equation}
\label{ConstraintEiTran}
    E_{i-tran}(t)<E_i(t)-(E_{min}+\delta_e), \forall i \in \mathcal{I}, t \in \mathcal{T}.
\end{equation}

\par Moreover, the terminals need to operate the daily tasks such as monitoring and sensing, which may consume some energy. According to~\cite{Chai2023}, this energy decreases linearly over time. As such, terminal battery energy decreases during daily operations at a rate of $\Delta E_1$ and increases during charging periods at a rate of $\Delta E_2$. Let $E_{min}$ be the minimum energy threshold for normal terminal operation, and then the energy of terminal $i$ at time slot $t+1$ can be expressed as follows:
\begin{equation}
E_i(t+1)=\operatorname{clip}\left(E_i\left(t\right)-\Delta E_1+\Delta E_2,E_{min},E_{max}\right),
\end{equation}

\noindent where the function $\operatorname{clip}(*_1, *_2, *_3)=\max(*_2, \min(*_1, *_3))$ ensures that terminal energy remains within the permissible range $[E_{min}, E_{max}]$.

%
%
\subsection{Problem Formulation}
\label{subsec:Problem Formulation}

\par The considered system concerns two objectives, which are minimizing total energy consumption while maximizing the average battery level of terminals, with an emphasis on ensuring equitable charging distribution among terminals with SWIPT. The key components of these objectives are analyzed as follows:
\begin{enumerate}
    \item Energy consumption is primarily attributed to the UAV propulsion energy $E^T_{u-move}$ and UAV communication energy $E^T_{u-tran}$, both of which are significantly influenced by flight trajectory and task offloading decisions.

    \item Enhancing the average battery energy of terminals requires optimization of the achievable EH rate $E_{u\rightarrow i}(t)$. According to Eq.~\eqref{E_u2i}, this rate is predominantly determined by the channel gain $h_i^t$ and antenna gain $G_i^t$, which can be optimized through controlling UAV position.
    
    \item Charging fairness among terminals necessitates careful consideration of the terminal charging sequence, which is directly related to flight trajectory planning of the UAV.
\end{enumerate}

\par To achieve these objectives, we aim to achieve our objectives by optimizing task offloading decisions ($\mathbf{O}$) and UAV trajectory planning, which is decomposed into velocity ($\mathbf{v}$) and direction ($\boldsymbol{\theta}$). Specifically, we jointly consider the following decision variables:
\begin{enumerate}
    \item $\mathbf{v}=\{v(t)\mid v(t)\in\mathcal{V}^t,t\in \mathcal{T}\}$, a vector consisting of continuous values that denotes the velocity of the UAV movement at each time slot.

    \item $\boldsymbol{\theta}=\{\theta(t)\mid\theta(t)\in\vartheta^t,t\in \mathcal{T}\}$, a vector consisting of continuous variables, denotes the angle of the UAV movement at each time slot.

    \item $\mathbf{O}=\{o_i^t\mid o_i^t\in \mathcal{O}^t, i\in \mathcal{I}, t\in \mathcal{T}\}$, a vector consisting of discrete variables, denotes the task offloading decisions of the terminals.
\end{enumerate}

\par Based on the above analyses, we aim to address the following optimization objectives simultaneously.

\par \textit{Optimization Objective 1:} Minimize the total system energy consumption, formulated as follows:
\begin{equation}
    E^T_{total} = E^T_{uav} + E^T_{terminal},
\end{equation}

\noindent where $E^T_{uav}$ and $E^T_{terminal}$ are calculated using Eq.~\eqref{Euav} and Eq.~\eqref{Eterminal}, respectively.

\par \textit{Optimization Objective 2:} Maximize average terminal battery energy while ensuring charging fairness in SWIPT, expressed as follows:
\begin{equation}
    F_{energy}(t)=J(t)\cdot\frac{\sum_{i=1}^I E_i(t)}I,
\end{equation}

\noindent where $J(t)$ represents Jain's fairness index~\cite{Jain1998}, quantifying the energy distribution among terminals:

\begin{equation}
    J(t)=\frac{\left(\sum_{i=1}^IE_i(t)\right)^2}{I\cdot\sum_{i=1}^I\left(E_i(t)\right)^2},
\end{equation}

\noindent where the $J(t)$ value approaching 1 indicates more equitable resource allocation across terminals.

\par In summary, considering the above objectives, we formulate the optimization problem as follows:
\begin{subequations}
\begin{align}
    \textbf{P}:\quad  \underset{\mathbf{v},\boldsymbol{\theta},\mathbf{O}}{\text{min}} &\sum_{t}^T E_{total}(t) + \underset{\mathbf{v},\boldsymbol{\theta},\mathbf{O}}{\text{max}} \sum_{t}^T F_{energy}(t) \label{P}\\
    \text{s.t.}\ \ 
    &p_u^0=(0,0),  \label{Pa} \\
    &0 < \eta\leq 1,  \label{Pb} \\
    &E_{min}\leq E_i(t)\leq E_{max}, \forall i\in\mathcal{I},t\in\mathcal{I},  \label{Pc} \\
    &\frac{\left(1 - o_i^t\right) C_iD^t_{i,p}}{f_i}+o_i^t(\tau_{up}+\tau_S)\leq\tau, \forall i\in\mathcal{I},t\in\mathcal{I},  \label{Pd} \\
    &R_{min}\leq\min\left\{R_{u\to i},R_{i\to u}\right\}, \forall i\in\mathcal{I},  \label{Pe} \\
    &0\leq v(t)\leq\ v_{\max}, \forall t\in\mathcal{T},  \label{Pf} \\
    &0\leq\theta(t)\leq2\pi, \forall t\in\mathcal{T},  \label{Pg} \\
    &C_iD^t_{i,p}\leq\tau f_i, \forall i\in\mathcal{I}, \ \label{Ph} \\
    &(\ref{ConstraintEiCom}),(\ref{ConstraintEiTran}).  \label{Pj}
\end{align}
\end{subequations}

\noindent where $v_\mathrm{max}$ represents the maximum flight velocity of the UAV, and $R_\mathrm{min}$ denotes the minimum required communication rate. The constraints ensure system feasibility: Constraint~\eqref{Pa} fixes the initial position of the UAV. Constraint~\eqref{Pb} restricts the power splitting ratio $\eta$ in SWIPT between 0 and 1. Constraint~\eqref{Pc} maintains terminal battery energy within normal bounds $[E_{min}, E_{max}]$. Constraint~\eqref{Pd} limits the combined local computation and task offloading time to within one time slot. Constraint~\eqref{Pe} guarantees that both uplink and downlink communication rates exceed the minimum threshold. Constraint~\eqref{Pf} and Constraint~\eqref{Pg} regulate the flight velocity and direction of the UAV. Constraint~\eqref{Ph} ensures terminals possess sufficient computational capacity to process tasks within the allocated time frame. Moreover, since optimization problem P consists of two optimization objectives with different units, where $E_{total}$ represents system energy consumption (J) and $F_{energy}$ denotes terminal battery energy ($\mu \text{J}$) weighted by the Jain fairness index, we employ a normalized reward function to address unit inconsistency. The optimization problem is further solved using a DRL-based approach.

%
%
\section{DRL-Based Method}
\label{sec:DRL-Based Method}

\par In this section, we propose a DRL-based offline method to solve the formulated optimization problem. We begin by discussing the motivations for using DRL. Next, we reformulate the optimization problem within an MDP framework. Finally, we detail our proposed method, emphasizing its specific enhancements and comparative advantages for the application domain.

%
%
\subsection{Motivations of Using DRL}
\label{subsec:Motivations of Using DRL}

\par The problem ($\textbf{P}$) exhibits three distinct properties. First, the scenario under consideration involves a hybrid action space for the UAV, comprising discrete variables (task offloading decisions) and continuous variables (velocity and direction). This hybrid nature renders the formulated problem non-convex. Second, it involves two conflicting objectives. Specifically, reducing $E_{total}(t)$ requires minimizing UAV communication and movement, which consequently decreases energy received at terminals; conversely, improving $F_{energy}(t)$ necessitates more frequent charging and increased flight distances of the UAV, thereby raising energy consumption. Third, the problem encompasses dynamics and uncertainties. The mobility of the UAV dynamically alters both the system network topology and the relative UAV-terminal distances, thereby affecting communication link quality. Additionally, uncertainty arises from the dynamic arrival of terminal tasks, as the UAV possesses limited knowledge of current terminal statuses and environmental conditions, preventing accurate prediction of future offloading task arrivals and processing requirements.

\par Therefore, the problem ($\textbf{P}$) constitutes a non-convex mixed-integer nonlinear programming problem with conflicting bi-objective optimization that incorporates dynamics and uncertainties. Consequently, traditional optimization methods prove unsuitable for two primary reasons. First, traditional static optimization methods, such as convex or non-convex optimization, struggle to effectively address the highly dynamic and unpredictable nature of this problem~\cite{Sun2024a}. Second, evolutionary computation algorithms (\textit{e.g.}, particle swarm optimization) tend to converge to local optima in complex environments, thereby resulting in suboptimal performance~\cite{Zeng2022}.

\par Consequently, we employ DRL-based methods to solve the optimization problem ($\textbf{P}$) due to their superior robustness and adaptability in dynamic and uncertain environments. This approach provides a more flexible and effective solution for UAV trajectory and resource scheduling.

%
%
\subsection{MDP Formulation}
\label{subsec:MDP Formulation}

\par To adapt our optimization problem to the DRL-based approach, we first introduce the MDP to model the decision-making process of the UAV (i.e., the agent)~\cite{Li2024a, Liu2024b}. Specifically, an MDP is typically defined by the quintuple $\langle\mathcal{S},\mathcal{A},\mathcal{P},\mathcal{R},\gamma\rangle $, where $\mathcal{S}$, $\mathcal{A}$, $\mathcal{P}= P(s'|s, a)$, $\mathcal{R} = r(a, s)$, and $\gamma\in[0, 1]$ denote state space, action space, state transition probability, reward function, and discount factor, respectively. In this quintuple, the emphasis is on the components $\mathcal{S}$, $\mathcal{A}$, and $\mathcal{R}$, which are critical to our implementation.

%
%
\par \textit{State Space:} We consider that both the UAV and all terminals employ precise global positioning systems for real-time position tracking. Additionally, since terminals remain stationary, state transitions are represented solely by UAV position changes. Excluding other unobservable factors (\textit{e.g.}, terminal battery levels), we define state $s_t$ as follows:
\begin{equation}
    s_t = \{x_u^t,y_u^t\},t\in\mathcal{T}.
\end{equation}

\noindent This two-dimensional continuous state space simplifies training while maintaining real-world applicability.

%
%
\par \textit{Action Space:} As outlined in the problem ($\textbf{P}$), our MDP action space corresponds directly to the decision variables. In each time slot, the UAV needs to determine the offloading decisions for all terminals ($\mathcal{O}^t=\{o_i^t\}$, $i\in\mathcal{I}$) and movement parameters (velocity $v^t$ and angle $\theta^t$). Correspondingly, the action $a_t$ is defined as follows:
\begin{equation}
    \label{action_space_1}
    a_{t}=\{\mathcal{O}^t,v^{t},\theta^{t}\},t\in\mathcal{T},
\end{equation}

\noindent This hybrid action space, comprising $\lvert \mathcal{I} \rvert$ discrete variables ($\mathcal{O}^t$) and two continuous variables ($v^t$ and $\theta^t$), substantially increases learning complexity and necessitates careful policy network design in our optimization algorithm.

\par According to~\cite{Sun2025a}, the flight action of the UAV is defined as velocity and angle $(v, \theta)$ (rather than Cartesian coordinates $(x, y, z)$), which effectively captures the temporal dynamics of UAV movement. This approach facilitates practical implementation as it aligns with real-world UAV control commands. Furthermore, this scheme simplifies the action space while enhancing learning efficiency.

%
%
\par \textit{Reward Function:} The reward function $r(t)$ is crucial for the convergence of DRL algorithms. To enhance stability and generalization capability, we carefully design the reward function $r(t)$ with three components, which are optimization objectives, constraints, and some behavioral incentives. In this study, we aim to minimize the total system energy consumption and maximize average terminal battery energy while ensuring charging fairness. Due to the unit inconsistency, we normalize and combine the two components into a single scalar reward through weighted summation. Correspondingly, $r(t)$ is defined as follows:
\begin{equation}
    r(t)=-\rho_1 E_{total}(t) + \rho_2 F_{energy} - \overline{R} + \rho_3 R_{w} + R_{char},
\end{equation}

\noindent where $\rho_1$ and $\rho_2$ are normalization parameters to ensure equivalent order of magnitude between the first and second terms, which correspond to the two distinct optimization objectives in Eq.~\eqref{P}. Furthermore, we guide the UAV's action by designing a three-component reward function, which includes an out-of-bound penalty ($\overline{R}$), terminal bias reward ($\rho_3 R_{w}$), and charging reward ($R_{char}$). These components are defined as follows.
\begin{enumerate}
    \item \textit{The out-of-bound penalty $\overline{R}$} is a positive constant that exceeds the typical reward magnitude within a single time slot, specifically designed to constrain movement within the permissible flight area.
    
    \item \textit{The terminal bias reward $\rho_3 R_{w}$} provides differential rewards for accessing terminals based on their spatial distribution, where $\rho_3\in[0,1]$ modulates the impact of this term, and $R_{w}$ is formulated as:
    \begin{equation}
        R_{w} = R_b\sum^I_{i=1}w_i,
    \end{equation}

    \noindent where $R_b$ denotes the baseline reward parameter, and $w_i\in[0,1]$ represents the accessibility challenge weight assigned to terminal $i$. Note that a terminal with greater accessibility challenges means that it is more difficult for the UAV to explore the terminal and therefore will have a larger weight $w_i$, thereby incentivizing the UAV to explore trajectories that incorporate these difficult-to-reach terminals.
    
    \item \textit{The charging reward $R_{char}$} primarily incentivizes the UAV to prioritize terminals with lower energy levels, thus promoting efficient resource allocation, defined as follows:
    \begin{small}
        \begin{equation}
        R_{char}=
            \begin{cases}
                0 &\text{if no communication,}\\
                \Delta E_{charging}(t) + C &\text{if communicating with} \\&\text{the lowest-energy terminal,}\\
                \Delta E_{charging}(t) &\text{otherwise.}
            \end{cases}
        \end{equation}
    \end{small}
    
    \noindent where $\Delta E_{charging}(t)$ denotes the energy increase of the target terminal during time slot $t$, and $C$ is a positive constant that provides additional reward for charging the lowest-energy terminal.
\end{enumerate}

\par After establishing this MDP framework, we subsequently introduce the standard SAC algorithm.

%
%
\subsection{SAC Algorithm}
\label{subsec:SAC Algorithm}

\par SAC is an off-policy actor-critic DRL algorithm based on the maximum entropy RL framework. Unlike on-policy methods, off-policy methods typically utilize experience replay buffers to enhance sampling efficiency, making it appropriate for our MDP with large variations in transitions. Compared to other common off-policy methods, the maximum entropy framework enables SAC to effectively handle high-dimensional, complex continuous action spaces while providing superior convergence and stability~\cite{Pu2021}.

\par Unlike other DRL methods, SAC incorporates two objectives: maximizing cumulative rewards and maintaining policy stochasticity. Specifically, the core innovation of this algorithm is the inclusion of an entropy regularization term in the objective function, enabling the agent to preserve exploration capability during convergence~\cite{Haarnoja2018}. The optimal policy objective with entropy can be formalized as follows:
\begin{equation}\pi^{*}=\arg\max_{\pi}\sum_{t}\mathbb{E}_{(s_{t},a_{t})\sim\rho_{\pi}}\left[r(s_{t},a_{t})+\alpha\mathcal{H}(\pi(\cdot|s_{t}))\right],
\end{equation}

\noindent where $\mathcal{H}(\pi(\cdot|\mathbf{s}_{t}))$ represents the policy entropy under state $s_t$, quantifying the randomness in action selection. The temperature parameter $\alpha$ functions as a regularization coefficient that balances entropy maximization against reward optimization, significantly affecting convergence performance. Moreover, the authors in~\cite{Haarnoja2018a} propose an automatic adjustment method for $\alpha$, with the corresponding loss function as follows:
\begin{equation}
\label{loss function of alpha} L(\alpha)=\mathbb{E}_{a_t\sim\pi_t}\begin{bmatrix}-\alpha\log\pi_t(a_t|s_t)-\alpha\bar{\mathcal{H}}\end{bmatrix}.
\end{equation}

\par However, the standard SAC algorithm faces several limitations in practical applications. First, its convergence stability is problematic. Standard SAC typically employs multi-layer perceptrons (MLPs) as function approximators~\cite{Sami2022, Zhang2023}, but these networks often exhibit unstable convergence and overfitting when handling MDPs with numerous parameters~\cite{Zhao2021}. Second, the hybrid action space in our MDP, which combines discrete and continuous actions, presents inherent difficulties for standard SAC implementation. Third, significant variability in immediate rewards impedes effective policy evaluation by neural networks. Based on these challenges, we aim to enhance the standard SAC algorithm to achieve faster and more stable convergence in complex dynamic environments.

%
%
\subsection{SAC-SK Framework}
\label{subsec:SAC-SK Framework}

\par In this part, we propose the SAC-SK algorithm, which integrates three enhancement modules: the action simplification mechanism, simple recurrent unit (SRU), and modified Kolmogorov-Arnold networks (KAN). SAC-SK optimizes the flight trajectories of the UAV through the optimization of a normalized scalar reward function, thereby addressing the formulated problem. We first detail each module individually, then examine the comprehensive architecture and operational principles of the SAC-SK framework.

\begin{figure*}[!hbt] 
	\centering
	\setlength{\abovecaptionskip}{2pt}%
	\setlength{\belowcaptionskip}{2pt}%
	\includegraphics[width =6.5 in]{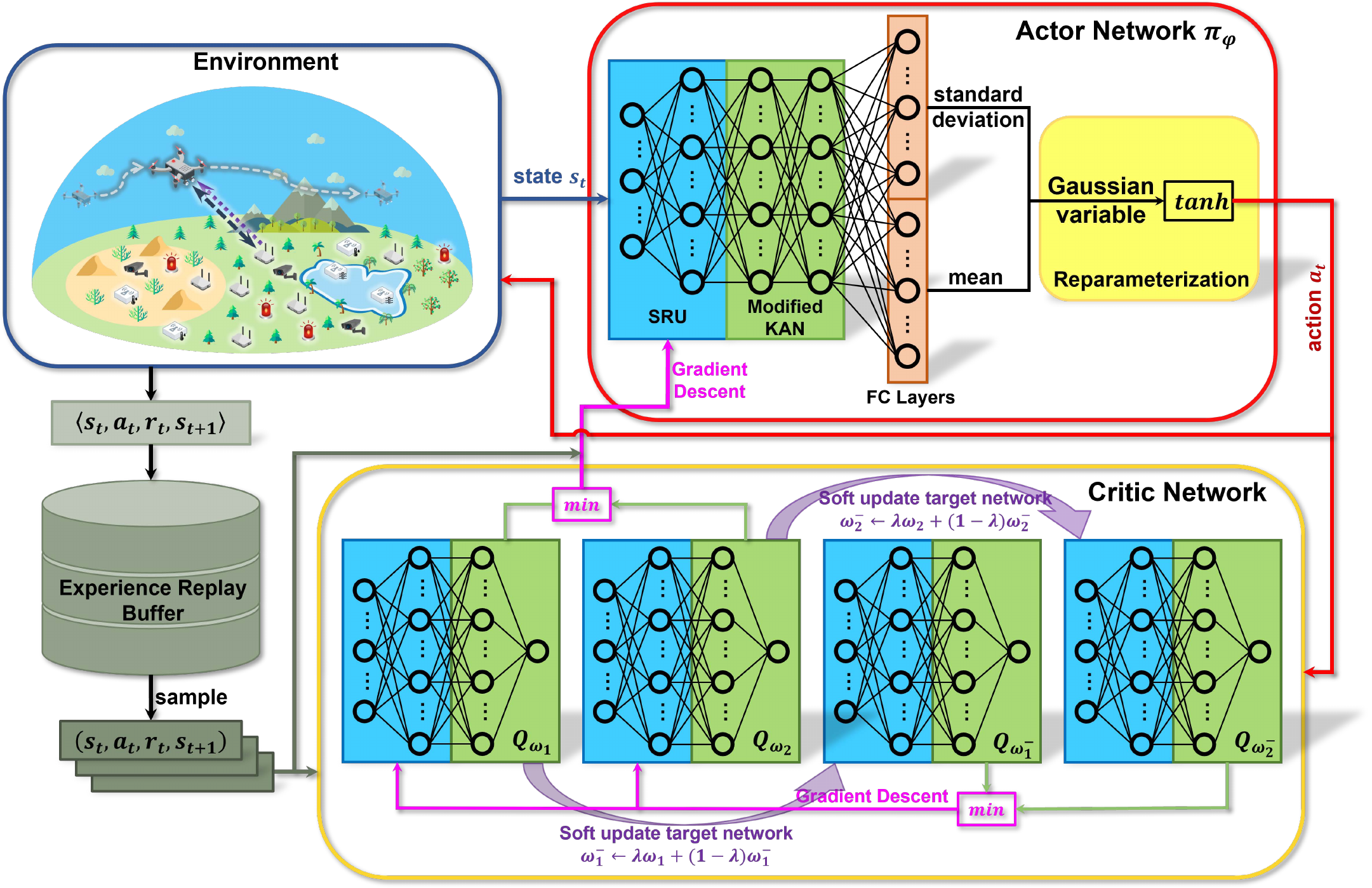}
    \caption{The architecture of SAC-SK, where the SRU layer and the modified KAN layer are integrated into the neural networks.}
    \label{SAC architecture}
	\vspace{-1em}
\end{figure*}

%
%
\subsubsection{Action Simplification Mechanism}
\label{subsubsec:Action Simplification Mechanism}

\par The standard SAC algorithm exhibits poor convergence and performance limitations when dealing with hybrid action spaces in our MDP framework. To address this issue, we propose an action simplification mechanism based on a greedy strategy. Specifically, through theoretical analysis of system operations, we develop a dynamic decision-making approach to replace original discrete decision variables, effectively transforming the hybrid action space into a continuous one.

\par To simplify the action space, we analyze the energy components $E_{i-tran}(t)$, $E_{u \rightarrow i}(t)$, and $E_{u-tran}(t)$ within a single time slot, as these variables directly correlate with the UAV-terminal distance. In contrast, $E_{u-move}(t)$ depends on both flight time and velocity~\cite{Sun2021}. With a constant volume of offloadable task data and communication distance $d_{u-i}$, the relationship between energy consumption and $d_{u-i}$ can be analyzed as follows:

\begin{enumerate}
    \item According to the first part of the Eq.~\eqref{Eterminal}, $E_{i-tran}(t)$ is proportional to the offloading time $\tau_{up}$, where $\tau_{up} \propto \frac{1}{R_{i \rightarrow u}}$. From Eqs.~\eqref{PathLoss},~\eqref{AntennaGain}, and~\eqref{R_i2u}, we establish that $R_{i \rightarrow u} \propto h_i^t G_i^t \propto \frac{1}{d_{u-i}}$. Thus, $E_{i-tran}(t) \propto d_{u-i}$.

    \item According to the second part of the Eq.~\eqref{Euav}, SWIPT divides $E_{u-tran}(t)$ into energy transmission and information transmission components. The total transmission time is determined by $\max{\left(\frac{1}{R_{u \rightarrow i}},\frac{1}{E_{u \rightarrow i}(t)}\right)}$, making $E_{u-tran}(t)$ proportional to this value. Based on Eqs.~\eqref{SNR},~\eqref{R_u2i}, and~\eqref{E_u2i}, both $R_{u \rightarrow i}$ and $E_{u \rightarrow i}(t)$ are proportional to $G_i^t$. Consequently, $E_{u-tran}(t) \propto d_{u-i}$ and $E_{u \rightarrow i}(t) \propto \frac{1}{d_{u-i}}$.
\end{enumerate}

\noindent Our analysis reveals that as the UAV approaches the terminal, both $E_{i-tran}(t)$ and $E_{u-tran}(t)$ decrease proportionally, while $E_{u \rightarrow i}(t)$ exhibits a significant increase. Additionally, $E_{u-move}(t)$ reaches its minimum value when the UAV maintains a constant flight velocity of 10 m/s~\cite{Yu2021}.

\par Therefore, under ideal conditions, maintaining a flight velocity of 10 m/s and communicating with the closest terminal minimizes $\sum_{t=1}^{T}E_{total}(t)$. However, this policy fails to guarantee maximization of $\sum_{t=1}^{T}F_{energy}(t)$, potentially creating energy replenishment imbalances among terminals. Consequently, the UAV needs to dynamically explore optimal flight velocities $v^t$ and angles $\theta^t$. Regarding offloading decisions, we simplify the process by directing communication exclusively to the nearest terminal with offloadable tasks, thereby reducing model complexity.  The action space is thus restructured as follows:
\begin{equation}
    a_{t}=\{v^{t},\theta^{t}\},t\in\mathcal{T}.
\end{equation}

\par This dimensional reduction to a two-dimensional continuous action space substantially improves learning efficiency.

%
%
\subsubsection{SRU}
\label{subsubsec:SRU}

\par Our MDP encompasses a large state space that incorporates environmental information and long-term dependencies, posing challenges for standard SAC algorithms to converge accurately and rapidly. To overcome this limitation, we introduce the SRU~\cite{Lei2018} for sequence modeling. SRU functions as a lightweight recurrent unit that delivers enhanced computational efficiency, improved scalability, and robust mathematical modeling while enabling high parallelization. In particular, the architectural framework of a single-layer SRU comprises two essential functional modules, which are a lightweight recurrence and a highway network.

\par
\begin{figure}[t]
    \centering
    \setlength{\abovecaptionskip}{2pt}%
    \setlength{\belowcaptionskip}{2pt}%
    \includegraphics[width =3.5in]{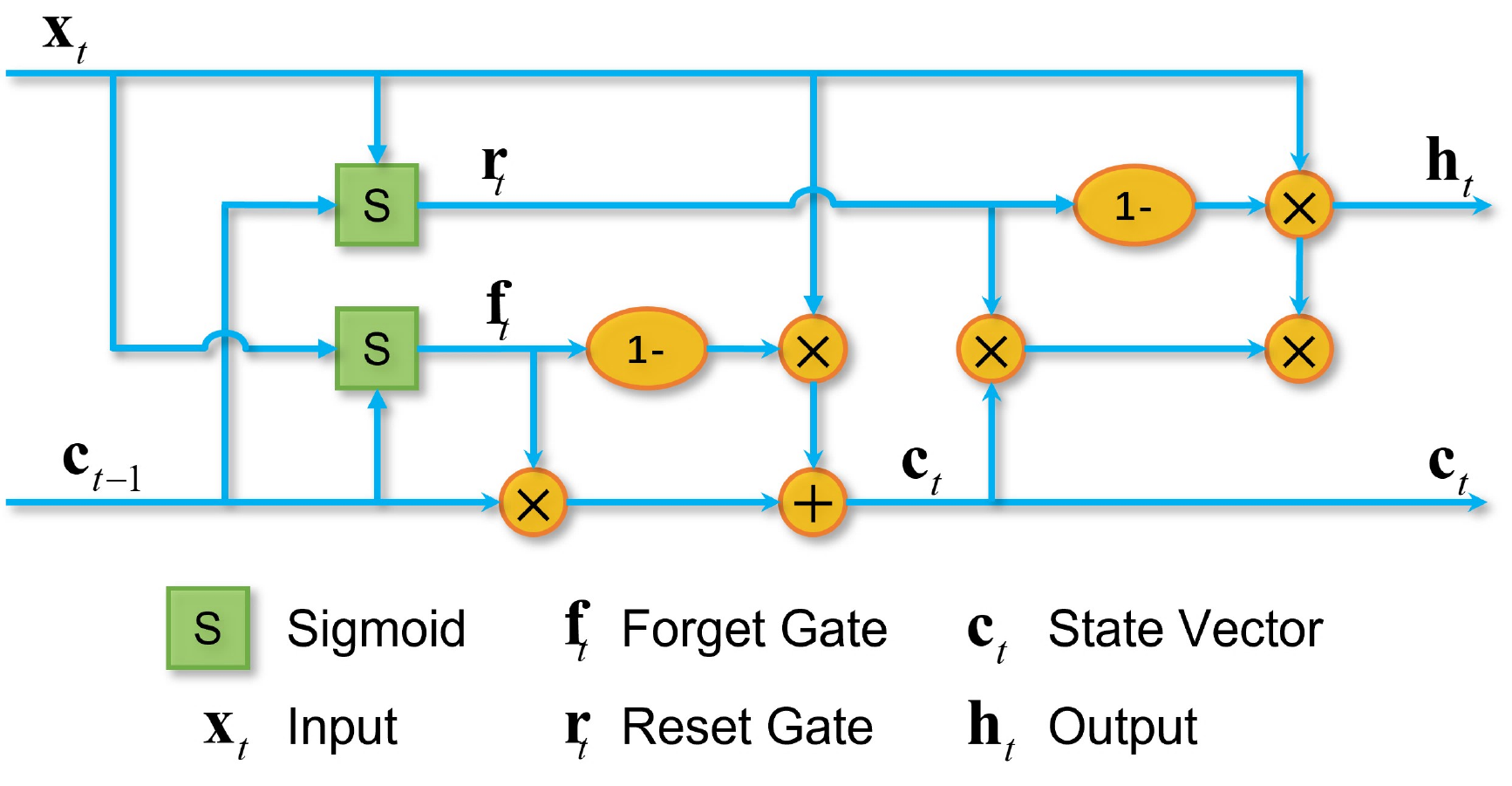}
    \caption{The schematic of SRU calculative architecture.}
    \label{SRU}
\end{figure}

 Compared to conventional recurrent architectures such as long short-term memory (LSTM)~\cite{Shi2015} and gated recurrent units (GRU)~\cite{Chung2014}, SRU strikes a balance between sequence dependence and independence. This design circumvents computational bottlenecks associated with temporal dependencies while enabling parallel computation, significantly enhancing processing speed and GPU resource utilization. As illustrated in Fig.~\ref{SRU}, SRU implements a forget gate $\mathbf{f}_t$ to regulate information flow, where the current state vector $\mathbf{c}_t$ is computed by integrating $\mathbf{f}_t$ with the previous state $\mathbf{c}_{t-1}$ and current input $\mathbf{x}_t$. Through this mechanism, SRU effectively preserves long-term dependencies in sequential data.

\par The integration of SRU yields a computational framework characterized by both structural elegance and representational power, while simultaneously offering exceptional scaling properties through enhanced parallel processing capabilities and optimized gradient transmission pathways. Moreover, SRU efficiently captures temporal dependencies and contextual information while significantly reducing computational complexity.

%
%
\subsubsection{Modified KAN}
\label{subsubsec:Modified KAN}

\par MLPs frequently struggle to effectively capture and extract latent information from complex sequential data, resulting in performance bottlenecks during challenging training scenarios. To overcome these limitations, we introduce the KAN~\cite{Liu2024a}, a promising neural architecture derived from the Kolmogorov-Arnold representation theorem that significantly enhances mathematical fitting accuracy. The architectural distinctiveness of the KAN framework manifests in its fundamental reformulation of inter-neuronal connections. Traditional fixed-coefficient parameters are replaced by dynamically learnable functions. Specifically, each traditional weight parameter at network intersections transitions to an adaptive univariate B-spline representation with comprehensive parameterization capabilities. Concurrently, computational nodes within this architecture implement straightforward signal aggregation without introducing additional transformation operations typically employed in traditional networks. This computational reorganization eliminates the rigid connectivity constraints characteristic of conventional neural architectures, instead enabling each pathway between processing units to exhibit autonomous adaptability. The resulting architecture demonstrates substantially enhanced capacity for nuanced information transmission control, thereby facilitating more sophisticated representational capabilities. Formally, a general $L$-layer KAN architecture can be expressed as:
\begin{equation}
    \begin{aligned}
        \mathrm{KAN}(\mathbf{x})= &\sum_{i_{L-1}=1}^{n_{L-1}}\phi_{L-1,i_{L},i_{L-1}}\\
        &\left(\cdots\left(\sum_{i_{1}=1}^{n_{1}}\phi_{1,i_{2},i_{1}}\left(\sum_{i_{0}=1}^{n_{0}}\phi_{0,i_{1},i_{0}}(x_{i_{0}})\right)\right)\right),
    \end{aligned}
\end{equation}

\noindent where $\phi_{l,j,i}$ represents the activation function connecting the $i$th neuron in layer $l$ to the $j$th neuron in layer $l+1$. These activation functions are constructed through a linear combination of basis functions $b(x)$ and spline functions, expressed as follows:
\begin{equation}
    \phi(x)=w_bb(x)+w_s\mathrm{spline}(x),
\end{equation}

\noindent where $b(x) = x /(1 + e^{-x})$ and $\mathrm{spline}(x) = \sum_i c_i B_i(x)$, with $w_b$, $w_s$, and $c_i$ serving as trainable parameters.

\par Note that the original KAN architecture demands substantial computational resources and employs complex network structures, thereby leading to prohibitive time complexity~\cite{Liu2024a}. To mitigate these issues, we propose a streamlined KAN variant that reduces hidden layer dimensionality and prunes redundant network layers, thereby achieving significantly lower computational overhead. Despite architectural simplifications, our modified KAN framework maintains superior mathematical fitting capabilities, primarily due to the integration of SRU components and the inherent advantages of the KAN paradigm.

%
%
\subsubsection{SAC-SK Architecture}
\label{subsubsec:SAC-SK Architecture}

\par In this part, we introduce the main architecture and procedures of SAC-SK.

\begin{algorithm}	
    \label{Algorithm 1}
    \SetAlgoLined
    \KwIn{The locations $\{p_i\}_{i\in \mathcal{I}}$ of terminals and the location $p_u^0$ of the UAV.}
    \KwOut{The velocity $\mathcal{V}$ and angle $\vartheta$ of the UAV in every time slot .}
    Initialize parameters of actor network $\varphi$ and soft Q networks $\omega_1$, $\omega_2$ randomly\;
    Initialize parameters of target Q networks: $\omega_1^- \leftarrow \omega_1$, $\omega_2^- \leftarrow \omega_2$\;
    Initialize experience replay buffer $\mathcal{B} = \varnothing$\;
    \For{iteration $m=1 \ to \ M$}
    {
        \If{Meet reset condition}{Reset environment with same random seed\;}
        \For{time slot $t=1 \ to \ T$}
        {
            Select action $ a_t \sim \pi_{\varphi}(s_t)$\;
            Get velocity $v_m(t)$ and angle $\theta_m(t)$ from $a_t$\;
            Execute action $a_t$\;
            Observe reward $r_t$ and next state $s_{t+1}$\;
            $\mathcal{B} \leftarrow \langle s_t, a_t, r_t, s_{t+1}\rangle$\;
            \If{Reach batch threshold}
            {
               Sample transitions $\{\langle s_b, a_b, r_b, s_{b+1} \rangle\}_{b=1,\dots,B_s}$ from $\mathcal{B}$\;
               Update soft Q network according to~\cite{Haarnoja2018a}\;
               Reparameterize and resample the new action\;
               Update actor network according to~\cite{Haarnoja2018a}\;
               Update temperature parameter $\alpha$\ according to Eq. \eqref{loss function of alpha}\;
               Soft update target Q network $\omega_i^- \leftarrow \lambda \omega_i + (1 - \lambda) \omega_i^-, i=1,2$\;
            }
            $\mathcal{V} \leftarrow v_m(t)$\;
            $\vartheta \leftarrow \theta_m(t)$\;
        }
    }
    \Return{$\mathcal{V},\vartheta$.}
    \caption{SAC-SK Algorithm}
    \vspace{-2pt}
\end{algorithm}

\par Nevertheless, certain distinctions exist between the actor network and the critic networks.

\par
\begin{figure}[t]
    \centering
    \subfigcapskip=-5pt 
	\subfigure[Actor Network]{
		\includegraphics[width=0.47\linewidth]{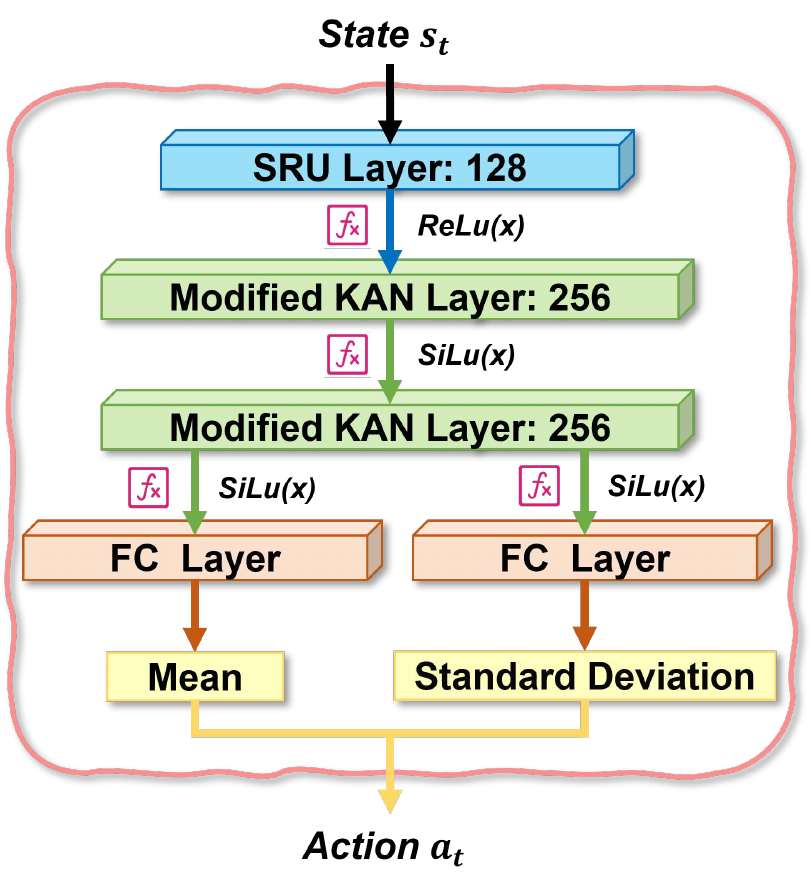}
        \label{Actor}
    }  
	\subfigure[Critic Network]{
		\includegraphics[width=0.46\linewidth]{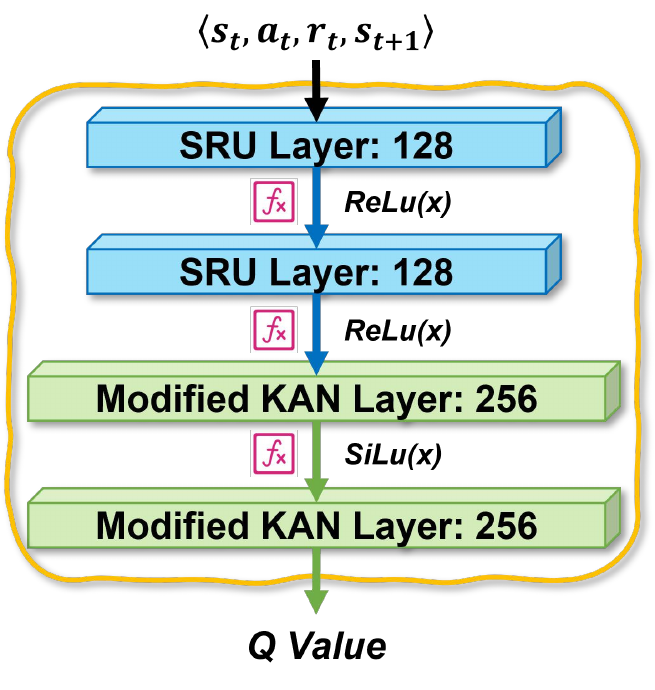}
        \label{Critic}
    }
    \caption{The neural network architecture of SAC-SK.}
    \label{neural networks}
\end{figure}

 As illustrated in Fig.~\ref{SAC architecture}, similar to conventional actor-critic frameworks, SAC-SK utilizes five enhanced neural networks, each integrating SRU layers and modified KAN layers. Specifically, the architecture processes input data (state $s_t$ or transition batches) through SRU layers that extract environmental information, maintain long-term dependencies, and generate a hidden information vector. This vector is subsequently processed by modified KAN layers, which leverage their mathematical representation capabilities for fitting. Nevertheless, the actor and critic networks differ in their output structures:
\begin{itemize}
    \item \textit{Actor Network}: As shown in Fig.~\ref{Actor}, two parallel fully connected layers (FC layers) process the modified KAN output to generate the mean and standard deviation of the action distribution, respectively.

    \item \textit{Critic Networks}: As shown in Fig.~\ref{Critic}, the modified KAN layers directly output a scalar value.
\end{itemize}

\par Algorithm~\ref{Algorithm 1} outlines the main steps of the SAC-SK algorithm. The procedure begins with network parameter initialization, followed by environment reset using consistent random seeds to ensure standardized initial conditions across episodes. During each time slot $t$, the algorithm: (1) selects action $a_t$ based on the current policy. (2) Executes the action to obtain transition $\langle s_t, a_t, r_t, s_{t+1} \rangle$. (3) Stores the transition in experience replay buffer $\mathcal{B}$. Once $\mathcal{B}$ reaches the predetermined threshold, SAC-SK samples transition batches for training and updates all neural networks along with the temperature parameter $\alpha$. This process iterates until reaching the preset time step limit, progressively approximating an optimal maximum entropy policy.

%
%
\subsubsection{Complexity Analyses}
\label{subsubsec:Complexity Analyses}

\par In this part, we analyze the computational and space complexity of the proposed SAC-SK algorithm in both the training phase and the execution phase.

\par \textbf{Training Phase:} The computational complexity of SAC-SK~\cite{Zhang2024} is $\mathcal{O}(\lvert \varphi\rvert+4\lvert \Omega\rvert+MTK+MT\lvert\varphi\rvert+MT(2\lvert\Omega\rvert)+MT/(4\lvert\Omega\rvert))$ in the training phase, which can be summarized as follows:

\begin{itemize}
    \item \textit{\textbf{Network Initialize:}} This phase involves the initialization of the neural network parameters. Specifically, the computational complexity is expressed as $\mathcal{O}(\lvert \varphi\rvert+4\lvert \Omega\rvert)$, where $\lvert \varphi\rvert$ denotes the number of parameters in the actor network, and $\lvert \Omega\rvert$ represents the average number of parameters in the four critic networks, which are $\lvert \omega_1\rvert$, $\lvert \omega_2\rvert$, $\lvert \omega_1^-\rvert$, and $\lvert \omega_2^-\rvert$, respectively, since these parameters are of the same order of magnitude.

    \item \textit{\textbf{Replay Buffer Collection:}} The complexity of collecting state transitions in the experience replay buffer is $\mathcal{O}(MTK)$, where $M$ is the number of training episodes, $T$ denotes the number of steps in an episode (i.e., the total number of time slots in the scenario), and $K$ represents the complexity of interacting with the environment.

    \item \textit{\textbf{Network Update:}} The updating phase is divided into three main parts, which are the updates of the actor network, the frequent updates of the soft Q networks, and the respective soft updates of target Q networks, respectively. Note that the temperature parameter $\alpha$ is a single parameter, so its update complexity is negligible. Therefore, the complexity for this phase is calculated as $\mathcal{O}(MT\lvert\varphi\rvert+MT(2\lvert\Omega\rvert)+MT/(4\lvert\Omega\rvert))$.
\end{itemize}

\par The space complexity of SAC-SK accounts for the storage of the neural network parameters and the transition structures $\langle s_t, a_t, r_t, s_{t+1} \rangle$ required to maintain the experience replay buffer $\mathcal{B}$. Therefore, the space complexity can be calculated as $\mathcal{O}(\lvert\varphi\rvert+4\lvert\Omega\rvert+\lvert\mathcal{B}\rvert(2\lvert s\rvert+\lvert a\rvert+1))$, where $\lvert \mathcal{B}\rvert$ represents the size of $\mathcal{B}$, while $\lvert s\rvert$ and $\lvert a\rvert$ represent the dimensions of the state space and action space, respectively.

\par \textbf{Execution Phase:} First, the computational complexity of SAC-SK is $\mathcal{O}(MT\lvert\varphi\rvert)$, resulting from the action selection based on the current state using the actor network. Second, the space complexity of SAC-SK is $\mathcal{O}(\lvert\varphi\rvert)$, primarily stemming from the parameters of the actor network.

\par 

\begin{figure*}[!hbt]
    \centering
    \subfigcapskip=-5pt 
    \subfigure[]{
        \includegraphics[width=0.23\linewidth]{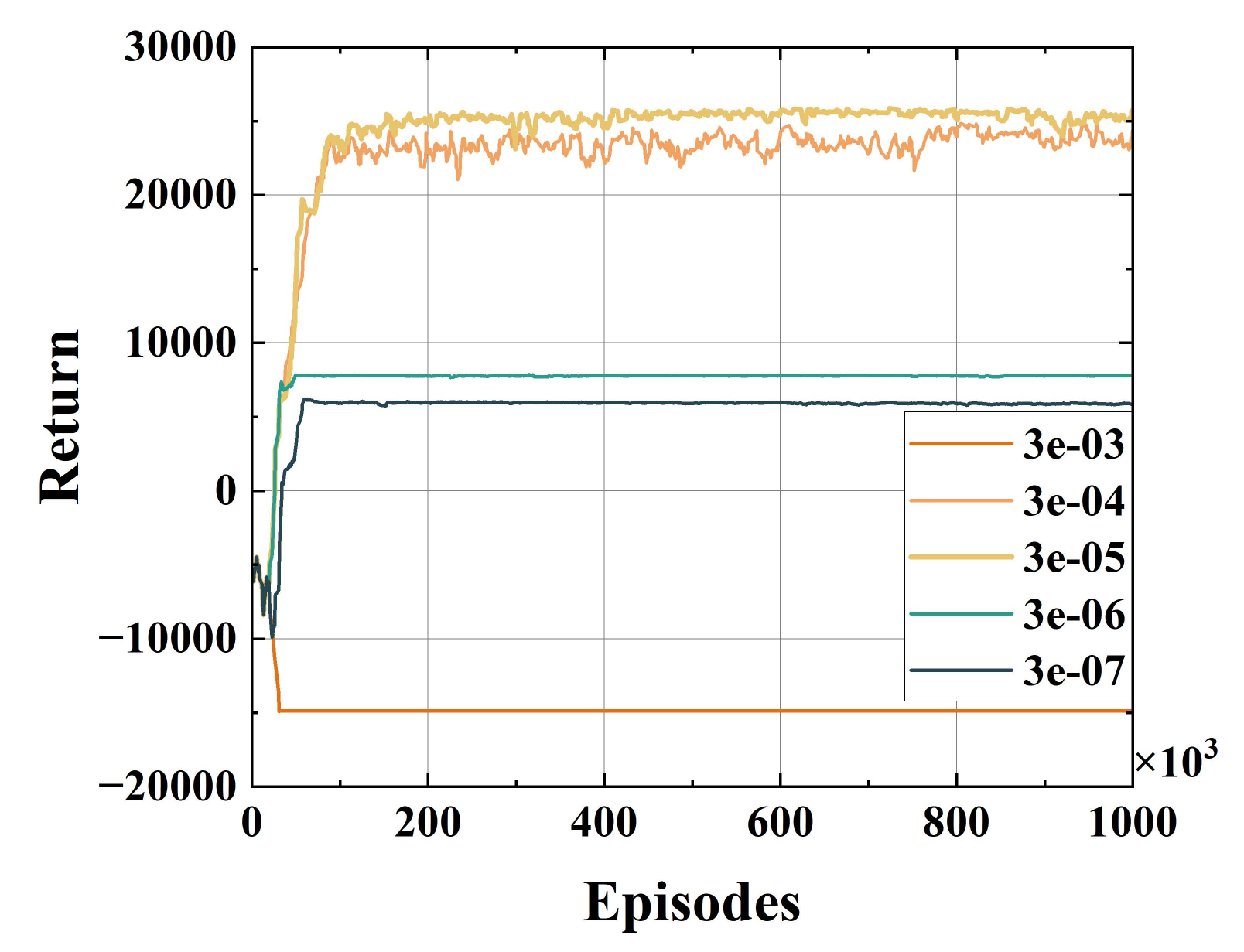}
        \label{p_lr}
    }  
    \subfigure[]{
        \includegraphics[width=0.23\linewidth]{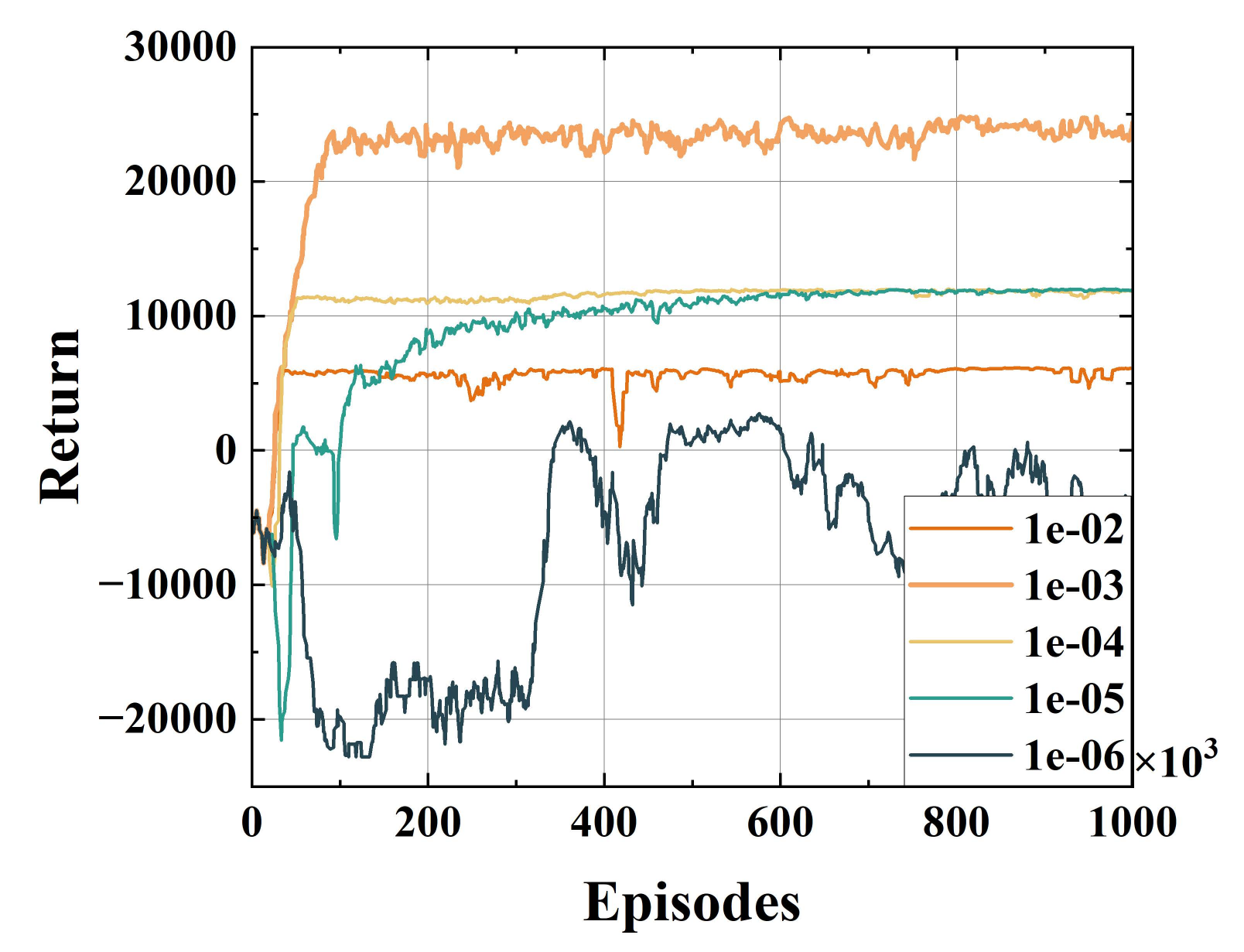}
        \label{q_lr}
    }
    \subfigure[]{
        \includegraphics[width=0.23\linewidth]{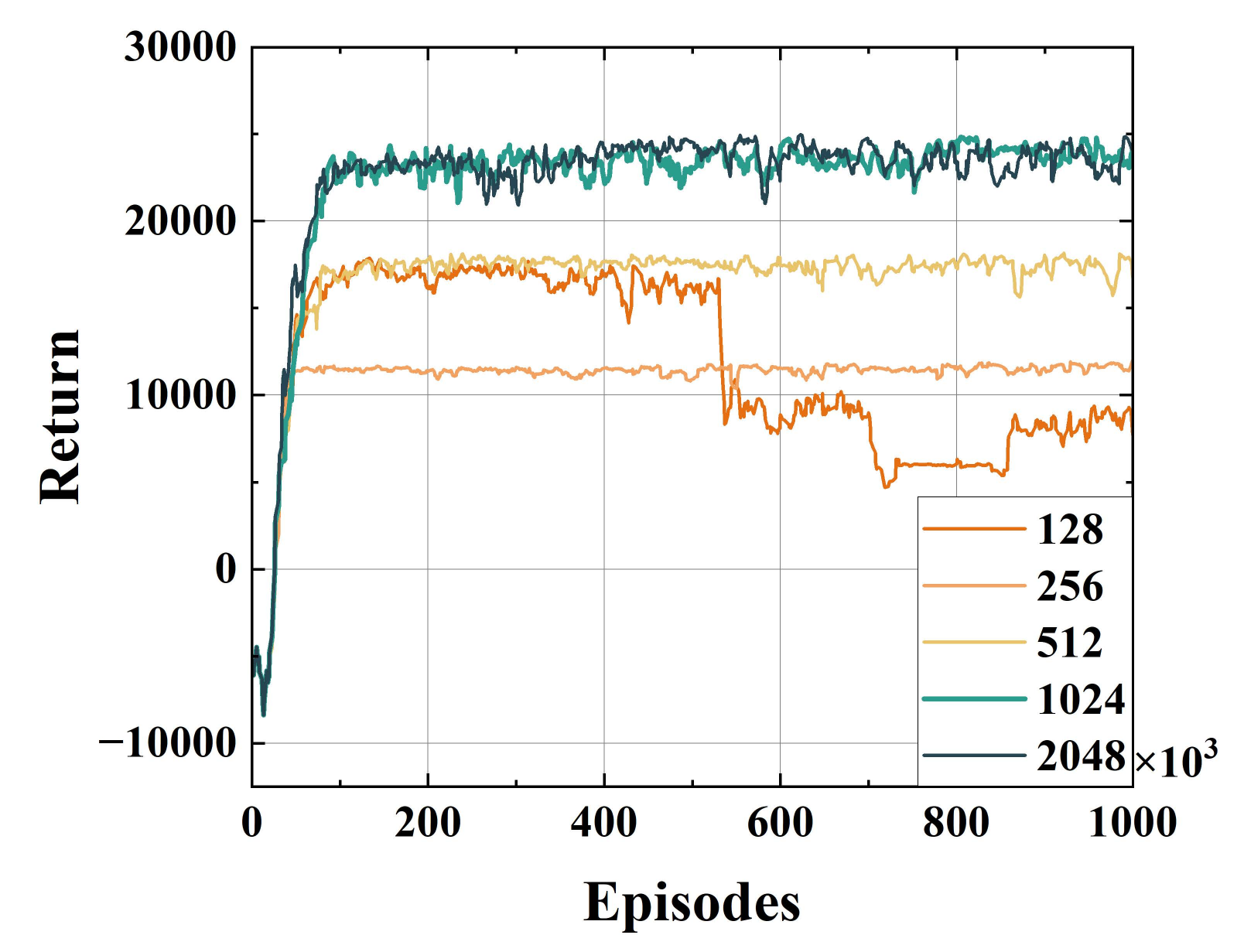}
        \label{batch_size}
    }
    \subfigure[]{
        \includegraphics[width=0.23\linewidth]{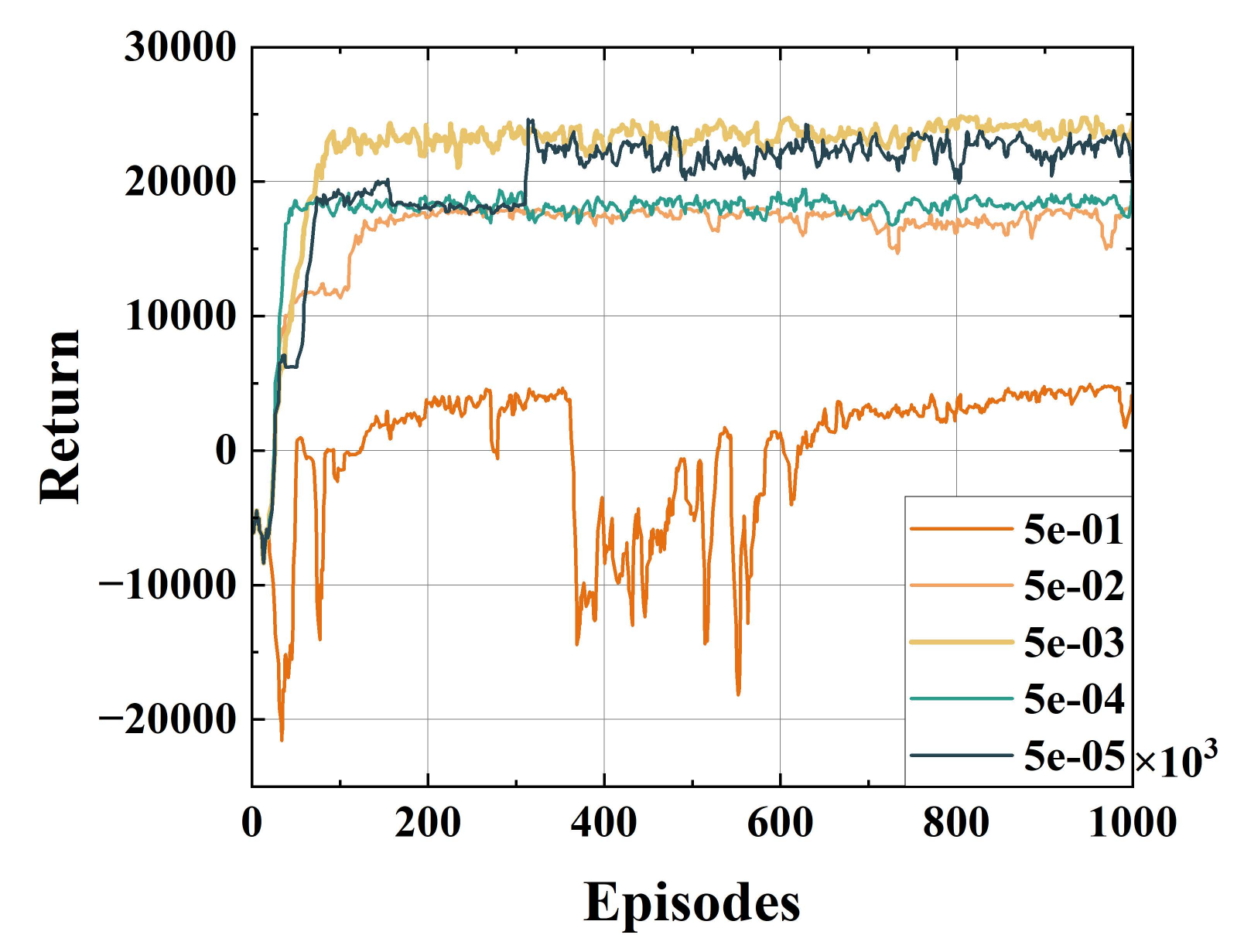}
        \label{smoothing}
    }
    \caption{Return under different key hyperparameters of SAC-SK. (a) The learning rate of the actor network $\lambda_{\varphi}$. (b) The learning rate of the critic network $\lambda_{\omega}$. (c) The batch size of the sample $B_s$. (d) target smoothing coefficient $\lambda$. }
    \label{hyperparameters}
    \vspace{-0.5em}
\end{figure*}

%
%
\subsection{Practical Implementation of SAC-SK-based UAV Management}
\label{subsec:Practical Implementation of SAC-SK-based UAV Management}

\par In this part, we elaborate on the application methodology of the proposed solution in practical scenarios, specifically divided into the training and deployment phase~\cite{Sun2025a}.

%
%
\subsubsection{Training Phase}
\label{subsubsec:Training Phase}

\par During the training phase, the neural network is initially trained in a simulation environment before deployment. Specifically, the proposed SAC-SK method does not depend on pre-collected real-world data. Instead, it interacts with a simulation environment constructed using mathematical models. Through this interaction, the SAC-SK algorithm gathers data to train the policy. We emphasize that our simulation is realistic, incorporating state-of-the-art mathematical models, including the air-to-ground channel model, non-linear EH model, energy consumption models for the UAV and terminals, and task computation models, among others. These models are derived from or based on real-world data, thereby ensuring the simulation environment closely reflects actual conditions. Consequently, the policy trained in this simulation can be effectively applied to real-world scenarios.

%
%
\subsubsection{Deployment Phase}
\label{subsubsec:Deployment Phase}

\par During the deployment phase, the pre-trained SAC-SK algorithm dynamically interacts with physical environments by processing real-time state data for decision generation. The SAC-SK strategies can be effectively implemented through the central control unit. Given that the UAV and terminals incorporate geolocation capabilities, the central control unit seamlessly obtains positional coordinates. During this phase, the SAC-SK eliminates the need for instantaneous calculations of transmission rates or energy expenditure metrics to determine reward values. Leveraging its acquired policy network, the system directly produces control commands that reflect predefined optimization objectives through existing spatial data (including device/UAV coordinates). The associated communication latency proves negligible since positional data transmission requires minimal bandwidth consumption.

%
%
\subsubsection{Environment Changes}
\label{subsubsec:Environment Changes}

\par When encountering environmental variations, pre-trained neural networks demonstrate adaptive potential through iterative parameter recalibration in modified environments. Specifically, operators may adjust critical simulation parameters (including UAV transmit power, terminal location distribution, etc.) to facilitate DRL model re-optimization. Advanced methodologies like transfer learning~\cite{Zhuang2021} significantly accelerate retraining convergence, while integrated architectural enhancements, particularly the action simplification mechanism, SRU layers, and modified KAN layers, collectively optimize learning efficiency. Building upon these foundations, the synergistic integration of edge computing with incremental learning paradigms enables seamless real-time updates for field-deployed network systems.

%
%
\section{Simulation and Analyses}
\label{sec:Simulation And Analyses}

\par In this section, we evaluate the performance of the proposed SAC-SK algorithm for solving the formulated optimization problem, with specific emphasis on convergence properties and computational efficiency.

%
%
\subsection{Simulation Setups}
\label{subsec:Simulation Setups}

\par In this part, we present the experimental setup, including scenario configuration, system parameters, baseline algorithms, and performance metrics.

%
%
\subsubsection{Simulation Setups}
\label{subsubsec:Scenario and Algorithm Setups}

\par In this work, we consider a scenario in which a UAV delivers both MEC and charging services to IoT terminals in a remote area. Specifically, the UAV operates at a fixed altitude of 10 m on a horizontal plane, with an initial position at the origin $(0,0)$ in Cartesian coordinates and a maximum velocity of 30 m/s. The service operation spans 30 seconds, discretized into 30 equal time slots.

\par Moreover, we consider a $40 \times 40\ \text{m}^2$ area without ground base stations, bounded by coordinates $\{(x, y) \mid x \in [-20, 20], y \in [-20, 20]\}$. Five IoT terminals are randomly distributed within this region using Poisson disk sampling to create a network topology centered around the origin. The defined boundary dimensions of 40 m provide sufficient space for UAV maneuverability during algorithm training. For larger geographical areas or more complex networks, clustering techniques such as k-means could partition the environment into multiple sub-regions, facilitating the deployment of multiple UAVs to execute our proposed method in each sub-network separately. Table~\ref{tab:SimulationSettings} provides all relevant simulation parameters.
    
\par For the proposed SAC-SK algorithm, the network architecture is illustrated in Fig.~\ref{neural networks}, with parameters optimized through extensive experimentation. Specifically, the actor network comprises a single-layer SRU with a hidden dimension of 128, coupled with the modified KAN layers and FC layers, each containing two layers with 256 neurons. The critic networks implement two-layer structures for both the SRU and modified KAN, each with 128 neurons. These architectural choices result from systematic parameter optimization to maximize performance and ensure convergence stability.

\begin{table}[!htbp]
\centering
\caption{Key simulation settings} 
\label{tab:SimulationSettings} 
\begin{tabular}{lp{3.5cm}l} 
\toprule 
\textbf{Symbol} & \textbf{Parameters} & \textbf{Values} \\
\midrule 
$f_c$ & Carrier frequency &  $2.4\ \text{GHz}$~\cite{ZhouHYS22}\\
$c$ & The speed of light & $3\times10^{8}\ \text{m/s}$~\cite{He2023}\\
$(\eta_{\mathrm{LoS}},\eta_{\mathrm{NLoS}})$ & Additional losses of LoS and NLoS links & $(0.1,21)$~\cite{Yaliniz2016}\\
$(a_1,b_1)$ & Parameters of LoS probability & $(4.88,0.43)$~\cite{Yaliniz2016}\\
$\sigma^2$ & Power of Gaussian white noise & $-174\ \text{dBm/Hz}$~\cite{Zhang2023a}\\
$P_{tran}$ & Transmit power of the UAV & $40\ \text{W}$~\cite{JiangXYFZL22}\\
$P_i$ & Transmit power of the terminals & $100 \ \text{mW}$~\cite{Wang2024a}\\
$B$ & Total bandwidth & $1\ \text{MHz}$~\cite{Yu2021}\\
$(a_2, b_2)$ & EH circuit parameters & $(150,0.014)$~\cite{Hua2022}\\
$P_{max}$ & The maximum received power of EH circuit & $24\ \text{mW}$~\cite{Hua2022}\\
$\eta$ & Power splitting ratio & $0.8$~\cite{ZhouHYS22,JiangXYFZL22}\\
$(k, \nu)$ & Effective capacitance coefficient & $(10^{-28},3)$~\cite{Jiang2020}\\
$\Delta E_1$ & Daily energy consumption for terminals & $50\ \mu\text{W}$~\cite{Chai2023}\\
$E_{max}$ & The maximum energy of terminal battery & $5\ \mu J$ \\
$E_{min}$ & The minimum energy of terminal battery & $800\ \mu J$ \\ 
$R_{min}$ & The minimum threshold of transmission rate & 22 Mbps \\
$D_{i,p}^t$ & Data size of the tasks &$10^3\ \text{bit}$~\cite{Wang2024a}\\
$C_i$ & The task computation density & $100\ \text{cycles/bit}$~\cite{ZhouHYS22,Sun2024}\\
$f_u$ & The CPU frequency of the UAV & $5\ \text{GHz}$~\cite{He2023}\\
$f_i$ & The CPU frequency of the terminals & $1\ \text{GHz}$~\cite{He2023}\\
$v_{max}$ & The maximum velocity of the UAV & $30\ \text{m/s}$~\cite{ZhouHYS22}\\
$I$ & Number of the terminals & $5$ \\
$H$ & Altitude of the UAV & $5\ \text{m}$~\cite{ZhouHYS22} \\
$T$ & Number of time slots & $30$ \\
$\tau$ & Length of one slot & $1\ \text{s}$~\cite{JiangXYFZL22,He2024}\\
- & The edge length of the area boundary  & $40\ \text{m}$\\
\midrule 
$(\rho_1,\rho_2)$ & Normalization parameters & $(0.3,1)$\\
$C$ & The charging reward constant & $300$\\
$\bar{R}$ & The out-of-bound penalty & $800$\\
$R_b$ & The baseline of terminal bias reward & 50 \\
$\alpha$ & Initial temperature parameter & $0.2$\\
$\gamma$ & Discount factor & $0.92$\\
$\lambda_{\varphi}$ & Learning rate of actor network & $3\times10^{-4}$\\
$\lambda_{\omega}$ & Learning rate of critic networks & $10^{-3}$\\
$B_s$ & The batch size of sample & $1024$\\
$\lambda$ & Target smoothing coefficient & $5\times10^{-3}$\\
- & Size of experience replay buffer & $10^{6}$\\
- & Policy frequency & $2$\\
- & Target network frequency & $1$\\
\bottomrule 
\end{tabular}
\end{table}

%
%
\subsubsection{Baselines Algorithms}
\label{subsubsec:Baselines Algorithms}

\par Performance evaluation of the proposed SAC-SK algorithm involves comparative analysis against several advanced DRL algorithms, including SAC, deep deterministic policy gradient (DDPG), twin delayed deep deterministic policy gradient (TD3), and proximal policy optimization (PPO). Specifically:
\begin{itemize}
    \item \textit{SAC}~\cite{Haarnoja2018a}: The original SAC algorithm utilizes MLP-based neural networks, which are the basis for improvements of SAC-SK. Similar to~\ref{subsubsec:Complexity Analyses}, the computational complexity of SAC is $\mathcal{O}(\lvert \varphi\rvert+4\lvert \Omega\rvert+MTK+MT\lvert\varphi\rvert+MT(2\lvert\Omega\rvert)+MT/(4\lvert\Omega\rvert))$.

    \item \textit{DDPG}~\cite{Lillicrap2015}: A classic off-policy algorithm designed specifically for continuous action space environments, known for its deterministic policy gradients. Similar to~\ref{subsubsec:Complexity Analyses}, the computational complexity of DDPG is $\mathcal{O}(2\lvert\varphi\rvert + 2\lvert\Omega \rvert + MTK + 2MT(\lvert \varphi\rvert + \lvert\Omega \rvert))$.

    \item \textit{TD3}~\cite{Fujimoto2018}: An enhanced variant of DDPG that addresses function approximation errors through twin delayed updates. Similar to~\ref{subsubsec:Complexity Analyses}, the computational complexity of TD3 is $\mathcal{O}(2\lvert\varphi\rvert + 4\lvert\Omega \rvert + MTK + \frac{3}{2}MT\lvert \varphi\rvert + 4MT\lvert\Omega \rvert)$.

    \item \textit{PPO}~\cite{Schulman2017}: A robust on-policy method that employs actor-critic architecture while maintaining computational efficiency. Similar to~\ref{subsubsec:Complexity Analyses}, the computational complexity of PPO is $\mathcal{O}(\lvert\varphi\rvert + \lvert\Omega \rvert + MTK + ME(\lvert \varphi\rvert + \lvert\Omega \rvert))$.
\end{itemize}

\par As can be seen, the computational complexity of the proposed SAC-SK is comparable to that of SAC, DDPG, TD3, and PPO. These methods share similar network architectures and training processes in deep reinforcement learning, thereby resulting in similar computational requirements.

\begin{figure}[!hbt]
    \centering
    \subfigcapskip=-5pt 
	\subfigure[]{
		\includegraphics[width =3.5in]{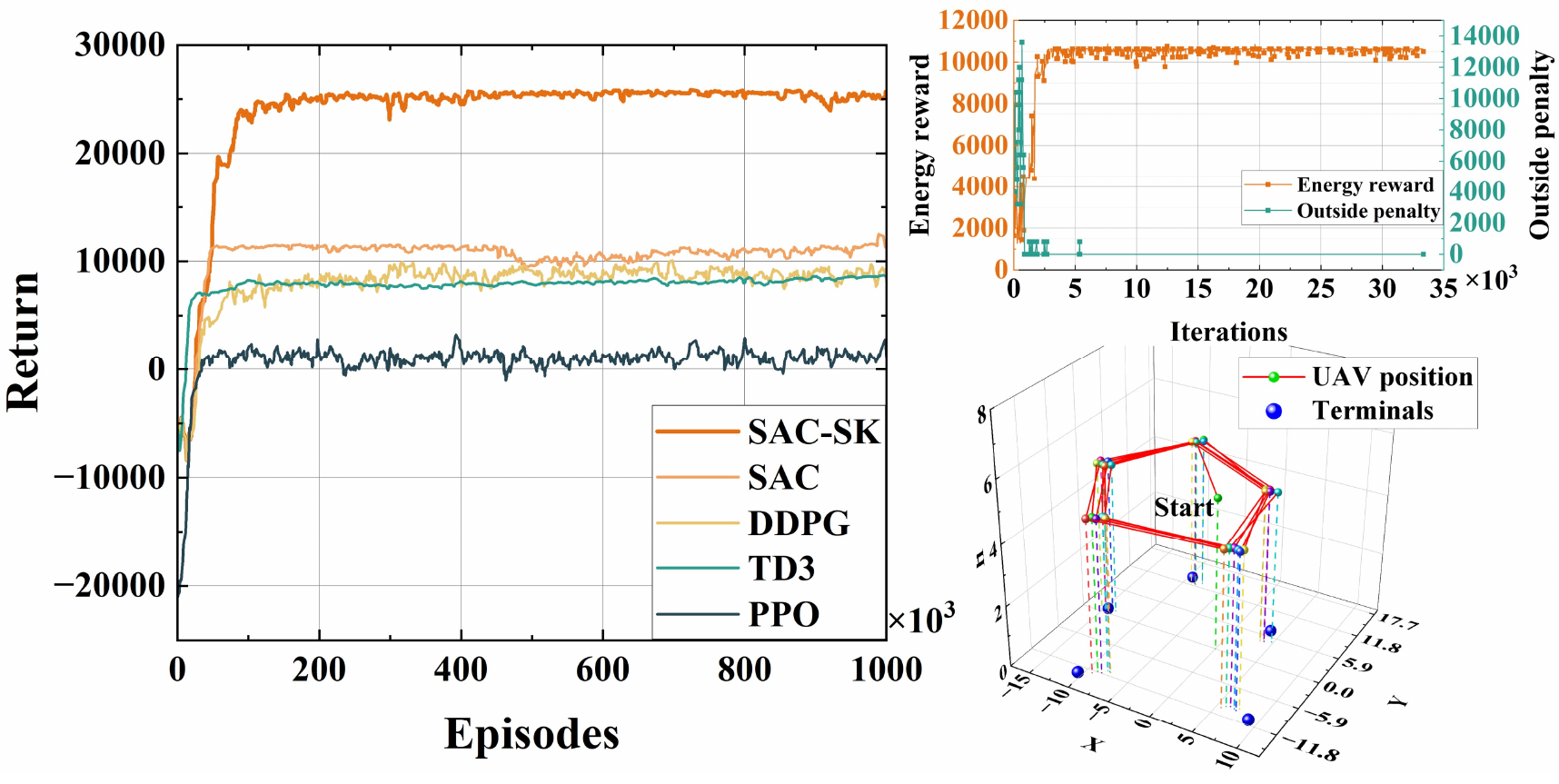}
        \label{result-1}
    }    
	\subfigure[]{
		\includegraphics[width =3.5in]{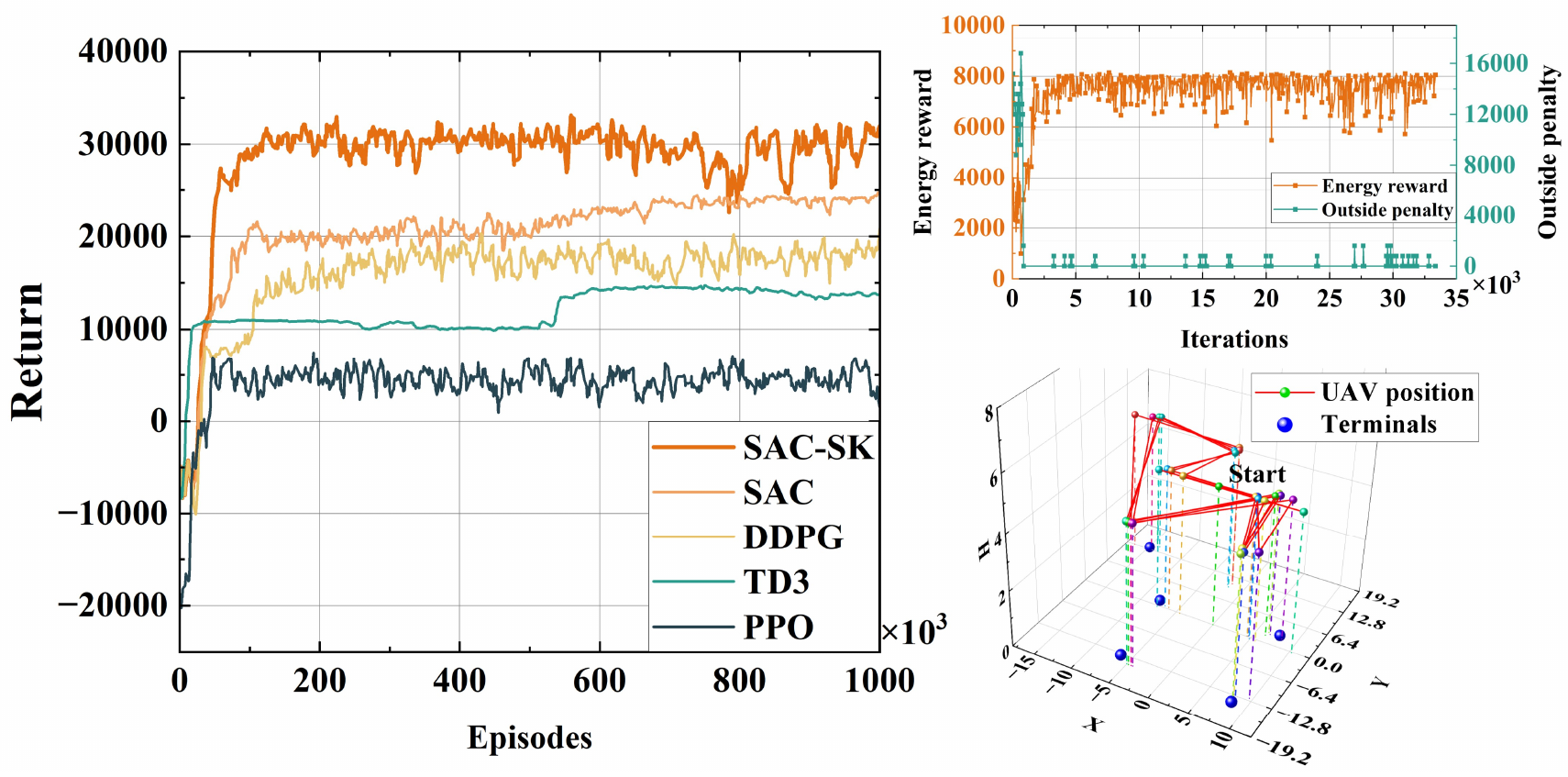}
        \label{result-3}
    }
    \subfigure[]{
		\includegraphics[width =3.5in]{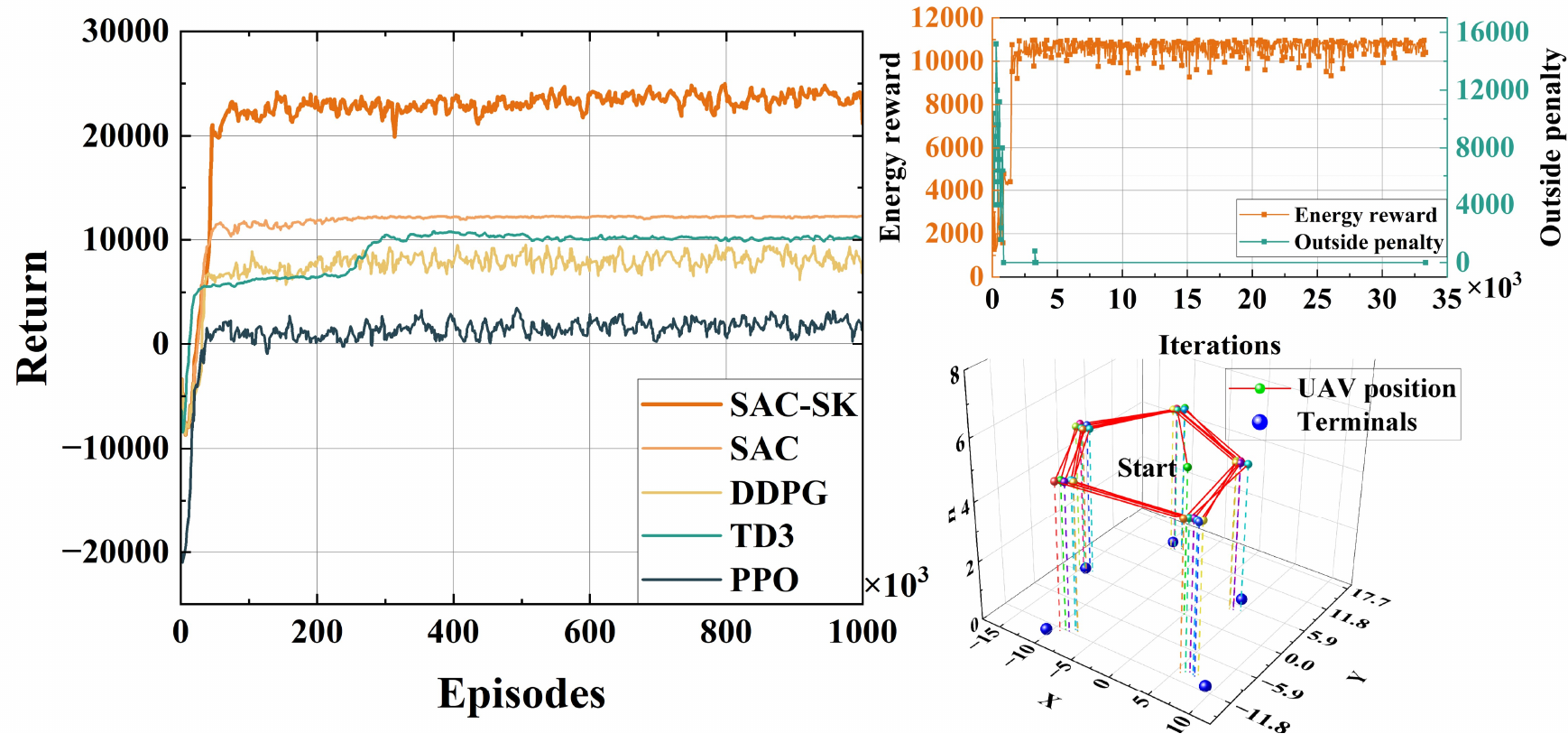}
        \label{result-5}
    }
    \caption{Simulation results under different random seeds, in which each subfigure includes a comparison between SAC-SK and baselines, the variation of the out-of-bound penalty $\bar{R}$ and the charging reward $R_{char}$ with iterations, and the trajectory of the UAV. (a) seed=1. (b) seed=3. (c) seed=5.}
    \label{result}
    \vspace{0em}
\end{figure}

%
%
\subsection{Performance Evaluation}
\label{subsec:Performance Evaluation}

\par In this part, we present a systematic performance evaluation of the proposed SAC-SK and analyze its operational characteristics.

%
%
\subsubsection{Convergence Evaluations}
\label{subsubsec:Convergence Evaluations}

\par To investigate the convergence and stability of SAC-SK, we conduct a comprehensive analysis involving the following key algorithmic parameters:
\begin{enumerate}
    \item \textit{Discount factor $\gamma$.} The experimental framework utilizes a time step of 30 slots in an iteration, thereby necessitating the consideration of rewards extending to at least the 30th future step. Applying the formulation $\gamma \approx 0.1^{1/t}$~\cite{Sutton1998} with $t=30$ yields a discount factor value of 0.92 approximately.

    \item \textit{Learning rate of actor network $\lambda_{\varphi}$.} Fig.~\ref{p_lr} illustrates the impact of varying $\lambda_{\varphi}$ values on convergence behavior. The results reveal that $\lambda_{\varphi} = 3\times10^{-5}$ yields optimal convergence performance, demonstrating improvements over alternative values.

    \item \textit{Learning rate of critic networks $\lambda_{\omega}$.} Fig.~\ref{q_lr} demonstrates that both excessively small and large $\lambda_{\omega}$ values adversely affect convergence. Although $\lambda_{\omega}=1\times10^{-4}$ facilitates rapid initial convergence, $\lambda_{\omega}=1\times10^{-3}$ exhibits superior long-term convergence performance. Therefore, we set $\lambda_{\omega}=1\times10^{-3}$ for optimal results.

    \item \textit{The batch size of sample $B_s$.} Fig.~\ref{batch_size} illustrates that while batch size exhibits minimal influence on convergence relative to other hyperparameters (with the exception of $B_s=128$), it substantially affects training stability. A batch size of $B_s=1024$ was selected to optimize convergence stability throughout the training process.

   \item \textit{Target smoothing coefficient $\lambda$.} Fig.~\ref{smoothing} demonstrates the influence of $\lambda$ on convergence. The result shows that SAC-SK achieves optimal convergence at $\lambda = 5\times10^{-3}$, a parameter that controls the soft update mechanism for target Q networks as expressed by $\omega_i^- \leftarrow \lambda \omega_i + (1 - \lambda) \omega_i^-, i=1,2$.
\end{enumerate}

\begin{figure*}[t]
    \centering
    \subfigcapskip=-5pt 
    \subfigure[]{
	\includegraphics[width=0.31\linewidth]{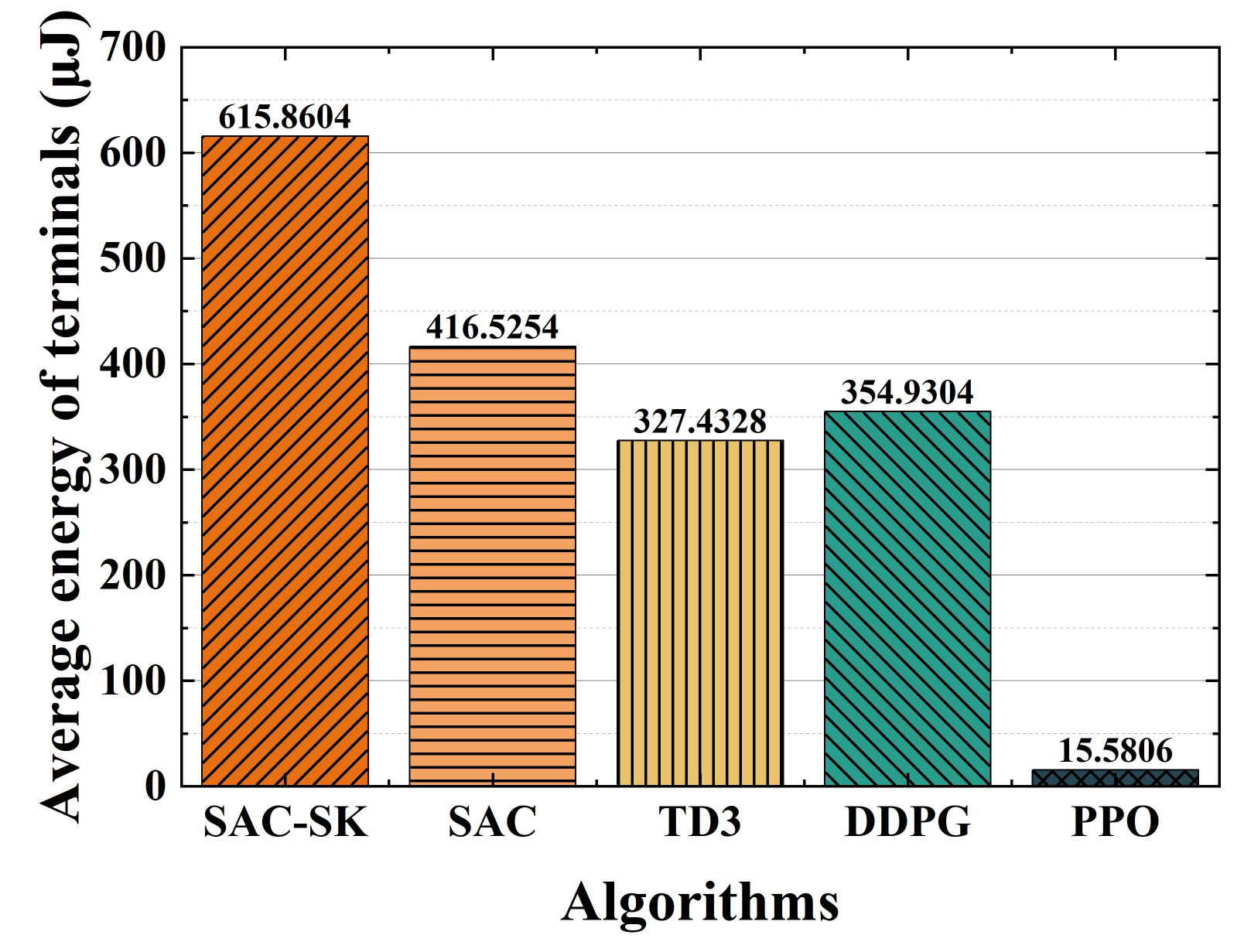}
        \label{AverageEnergy}
    }
    \subfigure[]{
	\includegraphics[width=0.31\linewidth]{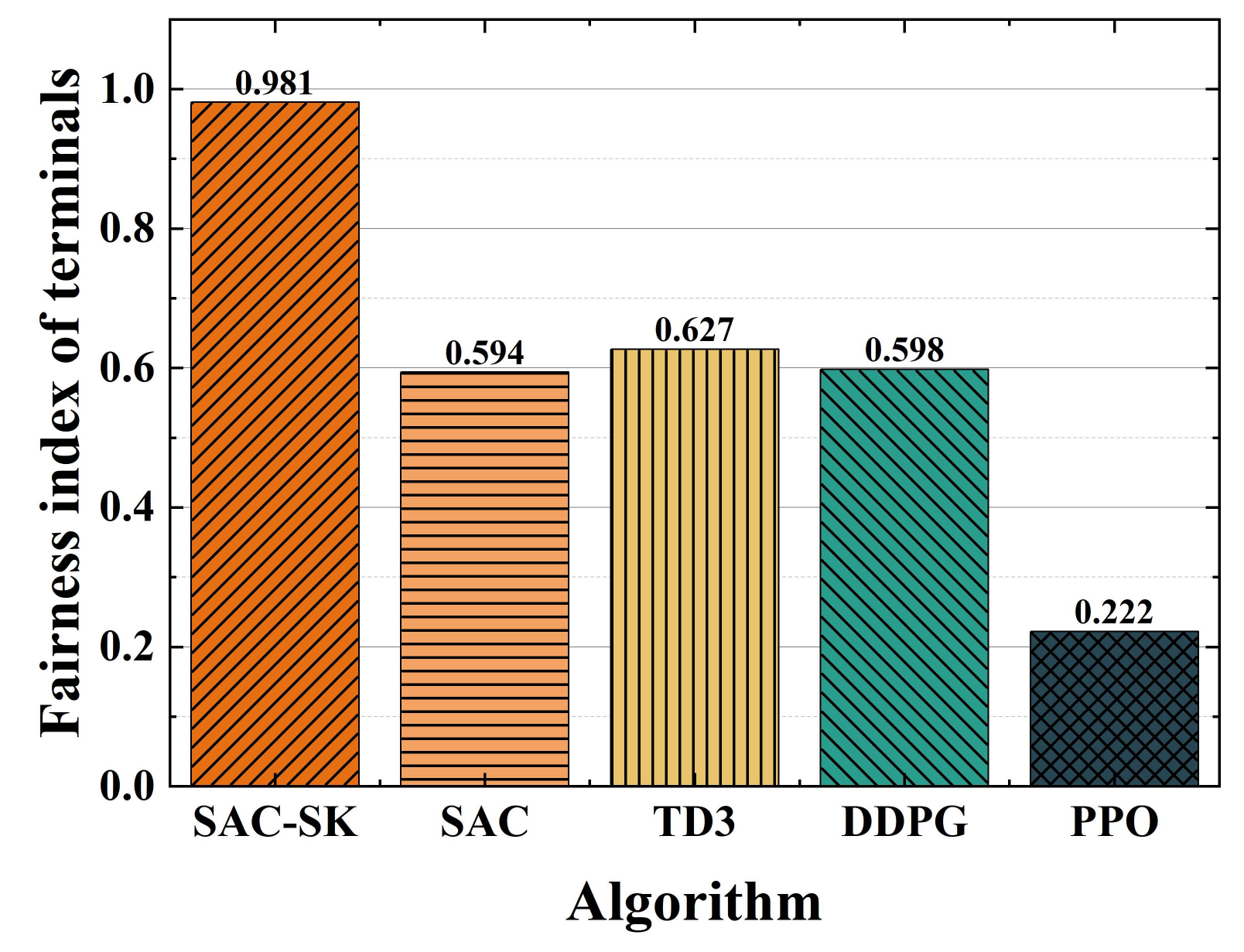}
        \label{Fairness}
    }
    \subfigure[]{
        \includegraphics[width=0.31\linewidth]{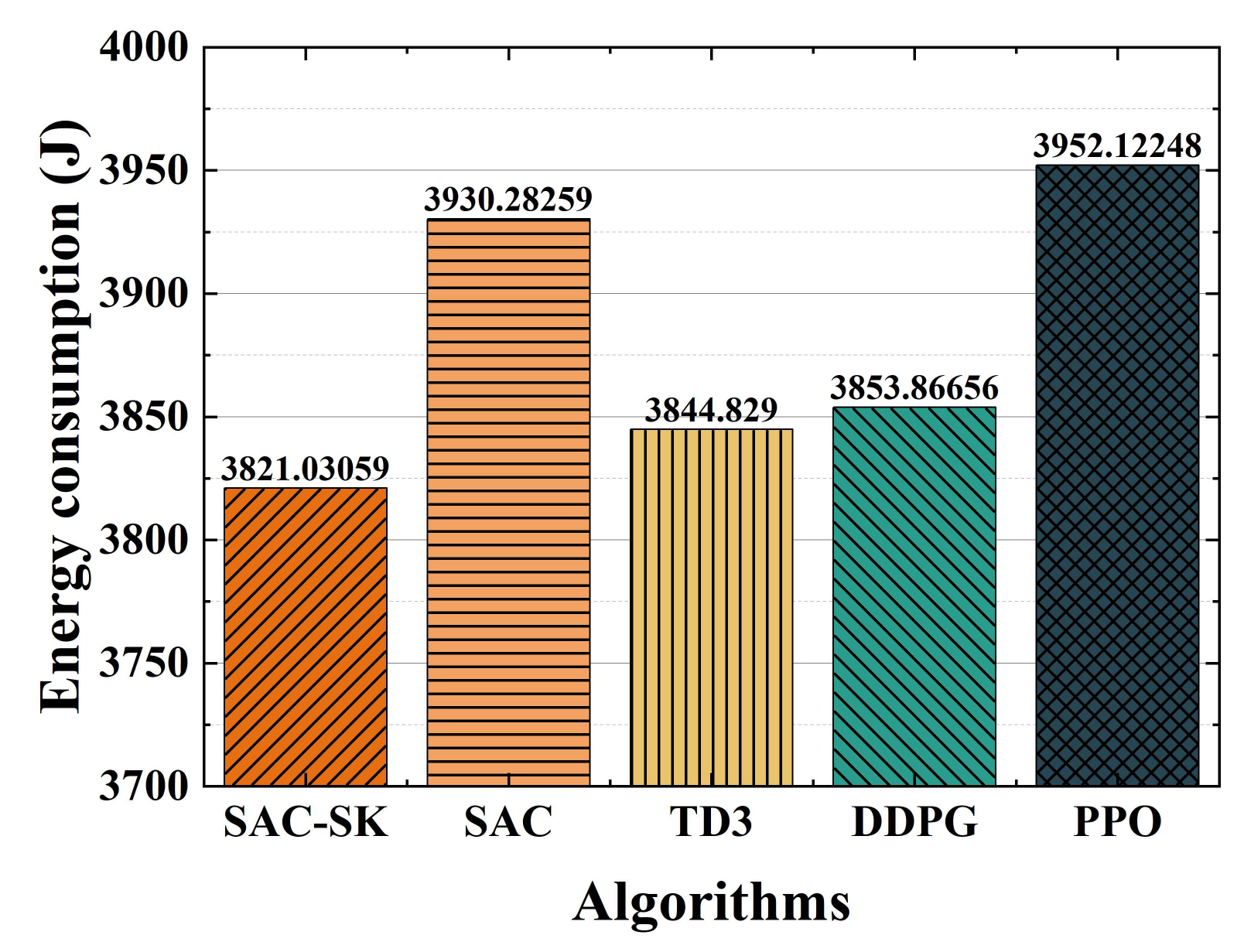}
        \label{EnergyConsumption}
    }  
    \caption{Comparison results of SAC-SK and DRL-based algorithms. (a) Average retained energy of terminals. (b) Charging fairness among terminals. (c) Total energy consumption of the system.}
    \label{ResultOfIndex}
\end{figure*}

\par In conclusion, we have identified the critical hyperparameters of SAC-SK and demonstrated its convergence and stability through rigorous experimental evaluation. Table~\ref{tab:SimulationSettings} summarizes the optimized hyperparameters derived from our systematic analysis.

%
%
\subsubsection{Comparisons with Different Baselines}
\label{subsubsec:Comparisons with Different Baselines}

\par In this part, we conduct some simulations of SAC-SK to assess its robustness and effectiveness and evaluate its performance through comprehensive comparisons with various baselines.

\par To verify the generalization capability of the proposed SAC-SK, we conduct a series of comparative simulations under distinct spatial distributions of terminals within the simulation area. Specifically, we consider three representative simulation scenarios by using different random seeds~\cite{Zhang2024}. These scenarios are shown in Figs.~\ref{result-1},~\ref{result-3} and~\ref{result-5}, and the details are as follows:
\begin{enumerate}
    \item The first case is uniform distribution. This scenario generates a relatively uniform spatial distribution of terminals, establishing a basic case for evaluating algorithm performance under standard conditions (Random seed=1 in this case). 
    \item The second case is a clustered distribution. This scenario features a high-density cluster of terminals with one outlier positioned at a significant distance from the main group. This configuration evaluates the convergence behavior of the algorithms to optimize trajectories when faced with potential local optima and coverage trade-offs (Random seed=3 in this case).
    \item The third case is asymmetric distribution. This scenario presents a terminal arrangement in which the terminal distribution center is significantly offset from the initial position of the UAV. This configuration evaluates the algorithms' exploration capabilities and adaptability to spatially biased configurations (Random seed=5 in this case).
\end{enumerate}

\noindent As illustrated in Fig.~\ref{result}, the proposed SAC-SK consistently outperforms all baselines across diverse random seeds, demonstrating robust generalization capability and efficient environmental information utilization.

\par In Fig.~\ref{ResultOfIndex}, we present three key performance metrics for the optimization objectives and compare SAC-SK against multiple baselines. These metrics are the average retained energy of terminals, charging fairness among terminals, and system energy consumption, respectively. As evidenced in Figs.~\ref{AverageEnergy} and~\ref{Fairness}, SAC-SK demonstrates substantial performance advantages over all baselines. Specifically, SAC-SK achieves 47.86\% higher average retained energy and 65.15\% improved charging fairness compared to standard SAC. These results confirm the effectiveness of SAC-SK in simultaneously maximizing terminal battery energy while ensuring fair charging distribution across all terminals. Furthermore, Fig.~\ref{EnergyConsumption} reveals that SAC-SK achieves the lowest system energy consumption among all tested algorithms, reducing consumption by 109.252 J relative to SAC. While TD3 and DDPG also exhibit lower energy consumption than SAC, the apparent efficiency of TD3 and DDPG is attributable to their suboptimal exploration patterns. The performance of TD3 and DDPG in inferior fairness indices (0.627 and 0.598, respectively) and average terminal energy levels (327.4328 $\mu \text{J}$ and 354.9304 $\mu \text{J}$, respectively) indicates convergence to local optima due to incomplete terminal coverage. In conclusion, SAC-SK successfully balances the inherent trade-off between the bi-objective optimization and delivering superior performance across all evaluation metrics.

\par Moreover, we conduct comprehensive performance analyses across all algorithms. The key findings are as follows:
\begin{enumerate}
    \item First, as an on-policy algorithm, PPO consistently underperforms compared to the off-policy algorithms across all random seeds. This performance gap is attributable to the experience replay mechanism in off-policy algorithms, which facilitates more efficient utilization of high-value transitions, thereby enhancing performance in complex environments.

    \item Second, SAC demonstrates superior convergence performance relative to DDPG and TD3, as evidenced by Figs.~\ref{result-1},~\ref{result-3}, and~\ref{result-5}. This result indicates that SAC provides a more effective foundation for algorithmic enhancement than other popular DRL algorithms. Specifically, the performance advantage derives from the maximum entropy RL framework, which simultaneously optimizes cumulative rewards and action randomness, thus improving both exploration efficiency and algorithmic robustness.

    \item Third, SAC-SK achieves superior convergence through its integration of SRU preprocessing for state-action inputs and modified KAN fitting for hidden information, thereby achieving the effectiveness of its action selection strategy. Comparative analysis across different random seeds confirms this finding. As illustrated in Fig.~\ref{result-3}, even in the most challenging scenario with complex terminal distribution (seed = 3), SAC-SK maintains a performance advantage with average return approximately 4000 values higher than SAC. This performance differential increases to over 10000 values in other scenarios, as evidenced by Figs.~\ref{result-1} and~\ref{result-5}. Simulation results demonstrate that integrating SRU as a temporal encoder effectively addresses the limitation of standard SAC in capturing temporal correlations in partially observable environments, while achieving preliminary modeling of environmental information and long-term dependencies with higher computational efficiency. Meanwhile, incorporating KAN as a function approximator effectively mitigates the deficiency of MLPs in standard SAC, where fixed activation patterns exhibit insufficient mathematical approximation capability for complex hidden information.

\end{enumerate}

\par Furthermore, we conduct a detailed analysis of the charging reward $R_{char}$ and the out-of-bound penalty $\bar{R}$. As shown in the upper right corners of Figs.~\ref{result-1},~\ref{result-3}, and~\ref{result-5}, the training process exhibits three distinct phases. In the initial phase (iterations 0 - 2000 approximately), the UAV learns boundary constraints, evidenced by $\bar{R}$ decreasing to zero. Correspondingly, random exploration of the UAV yields communication links with terminals, thereby causing rapid increases in $R_{char}$. During the intermediate phase (iterations 2000 - 5000 approximately), the UAV systematically explores the valid operational area, acquires terminal distribution information, and develops long-term memory representations. Concurrently, the experience replay buffer reaches its training threshold. This phase demonstrates reduced $R_{char}$ growth rate due to the increased complexity of reward maximization. The final phase (beyond 5000 iterations) achieves convergence stability while maintaining exploration of high-entropy strategies. These observations confirm that the designed reward function $r(t)$ effectively guides SAC-SK toward convergence.

\par Additionally, we visualize the flight trajectory of the UAV and its spatial relationships with ground terminals through a 3D format to present the results more intuitively.

%
%
\section{Conclusion}
\label{sec:Conclusion}

\par In this paper, we have investigated a directional antenna-enhanced UAV-assisted multi-terminal SWIPT-MEC system operating in infrastructure-free environments. Specifically, we have proposed a novel architecture wherein a UAV functions as both a base station and MEC server to provide charging and computational offloading services for energy-constrained ground terminals. In this system, we have formulated a bi-objective optimization problem to minimize system energy consumption and simultaneously maximize terminal battery energy while ensuring charging fairness among terminals. Subsequently, we have reformulated the original problem into an MDP to enhance computational tractability and system scalability. To address this MDP, we have proposed the SAC-SK algorithm to learn a maximum entropy optimal policy, thereby efficiently scheduling offloading decisions and UAV trajectory planning. Simulation results have demonstrated that SAC-SK significantly outperforms baselines across multiple performance metrics, while exhibiting robust generalization capabilities in diverse scenarios. This study has several limitations that are worth considering. First, static ground terminals may not fully capture the dynamics of real-world mobile scenarios. Second, the energy consumption model does not account for potential signal interference in dense deployment environments. Finally, although SAC-SK optimizes the time complexity and computational load during the training process, it still requires non-negligible computational resources, which may limit its practical deployment in resource-constrained UAV platforms. Future work will be extended along four dimensions, which include dynamic user mobility modeling, multi-UAV coordination, more lightweight neural network architectures,  and the integration of digital twin technology and blockchain-based solutions.

\normalem
\bibliographystyle{IEEEtran}
\bibliography{ref}
\end{document}